\documentclass[letterpaper,journal]{IEEEtran}
\usepackage{amsmath,amsfonts}
\usepackage{algorithmic}
\usepackage{algorithm}
\usepackage{array}
\usepackage[caption=false,font=normalsize,labelfont=sf,textfont=sf]{subfig}
\usepackage{textcomp}
\usepackage{stfloats}
\usepackage{url}
\usepackage{verbatim}
\usepackage{graphicx}
\usepackage{cite}
\usepackage{booktabs}   
\usepackage{multirow}   
\usepackage{graphicx}   
\usepackage{amsmath}    
\usepackage{amssymb}    
\usepackage{textcomp}   
\usepackage{enumitem}
\usepackage{marvosym}
\usepackage{pifont}
\usepackage{gensymb}
\usepackage{hyperref}
\usepackage{xcolor}

\makeatletter
\let\DexRoamOriginalMakeCaption\@makecaption
\long\def\@makecaption#1#2{%
  \ifx\@captype\@IEEEtablestring
    \footnotesize\bgroup
      \par\@IEEEtabletopskipstrut\noindent
      {\normalfont\footnotesize #1:\nobreakspace #2}%
      \par\addvspace{0.5\baselineskip}
    \egroup
    \@IEEEtablecaptionsepspace
  \else
    \DexRoamOriginalMakeCaption{#1}{#2}%
  \fi}
\makeatother

\definecolor{linkblue}{RGB}{18, 98, 243}

\begin{document}

\title{\textbf{DexRoam: Learning Mobile Bimanual Dexterous Manipulation from Egocentric Whole-Body Human Demonstrations}}

\author{
\IEEEauthorblockN{
\textbf{
Rui Zhou$^{1,2,*}$,
Yibo Yuan$^{4,2,*}$,
Junkai Zhao$^{2,*, \dagger}$,
Fangyuan Zhao$^{3}$,\\
Xiaoguang Zhao$^{5}$,
Shanghang Zhang$^{3,\textsuperscript{\Letter}}$,
Sirui Han$^{1,\textsuperscript{\Letter}}$
}
}\\
\IEEEauthorblockA{
$^{1}$The Hong Kong University of Science and Technology
$^{2}$Beijing Academy of Artificial Intelligence\\
$^{3}$State Key Laboratory of Multimedia Information Processing,
School of Computer Science, Peking University\\
$^{4}$Beihang University
$^{5}$Institute of Automation, Chinese Academy of Sciences\\
$^{*}$Equal contribution 
$\dagger$Project Leader 
\textsuperscript{\Letter}Corresponding author\\
Project Page: \textbf{\href{https://dexroam.github.io/}{\textcolor{linkblue}{https://dexroam.github.io/}}}
}
}

\markboth{}%
{Shell \MakeLowercase{\textit{et al.}}: A Sample Article Using IEEEtran.cls for IEEE Journals}

\IEEEaftertitletext{%
\begin{center}
\vspace{-1cm}
\vspace{-0.5\baselineskip}

\includegraphics[width=0.94\textwidth]{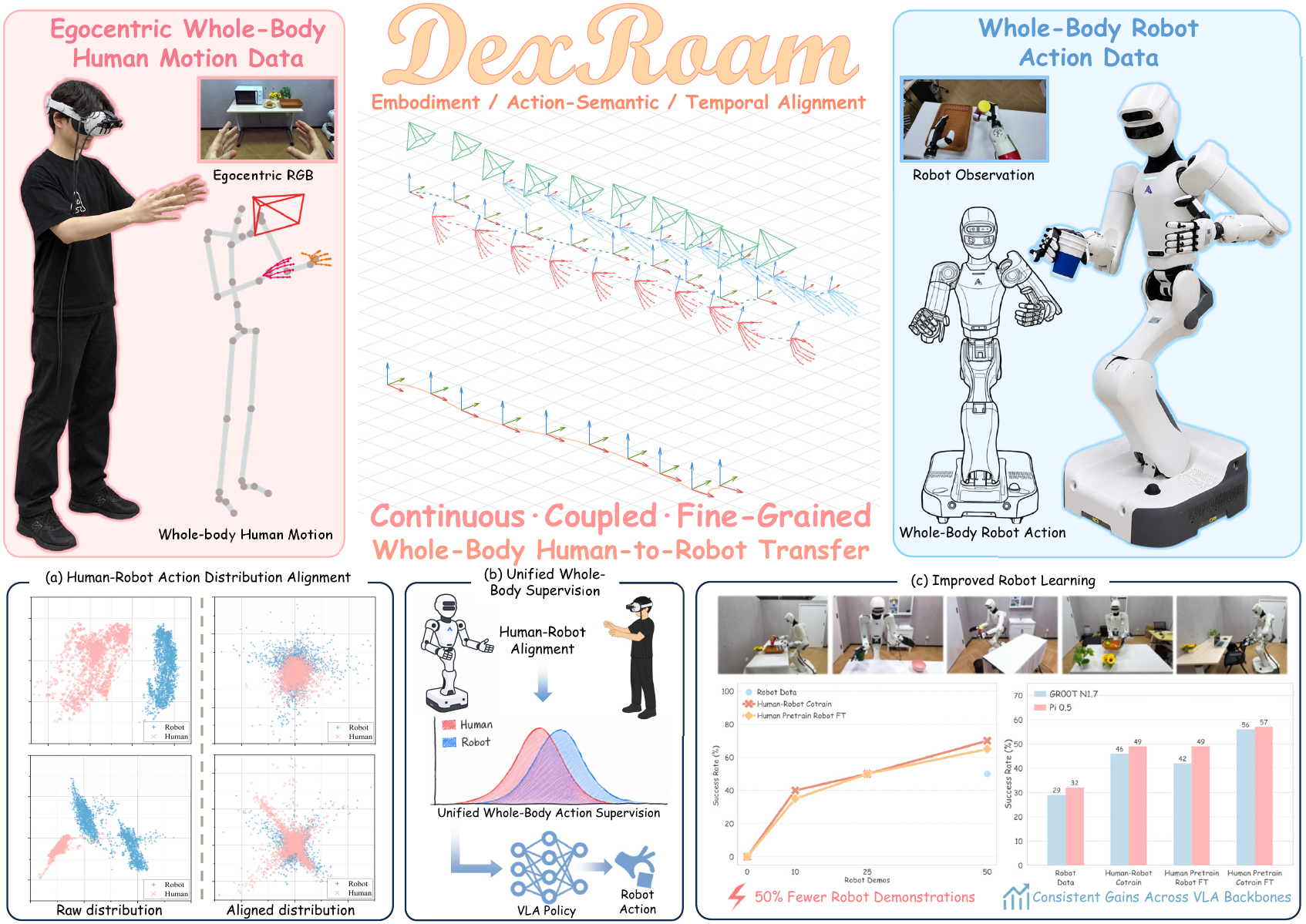}

\vspace{0.2\baselineskip}

\refstepcounter{figure}
\label{fig:teaser}

\parbox{0.94\textwidth}{
\small
\textbf{Figure \thefigure. Overview of the DexRoam framework.}
DexRoam unlocks motion-level human data for mobile bimanual dexterous manipulation by transforming egocentric whole-body demonstrations into a continuous, coupled, robot-compatible action space. DexRoam provides a complete pipeline for tracker-free human motion capture, human-to-robot alignment, and VLA-based policy learning, enabling scalable learning from human demonstrations.
}
\vspace{-0.2cm}
\vspace{0.2\baselineskip}

\end{center}
}

\maketitle

\begin{abstract}
Mobile bimanual dexterous manipulation requires continuous coordination of locomotion, whole-body motion, and finger-level dexterity within a single trajectory, creating a severe robot demonstration bottleneck. Egocentric human demonstrations offer a scalable alternative, but prior approaches ease the transfer by simplifying human motion, discarding exactly the fine-grained, coupled structure such tasks depend on. We present DexRoam, a complete system for learning mobile bimanual dexterous manipulation from human demonstrations, in which whole-body motion remains continuous and coupled throughout the human-to-robot transfer process. To enable scalable collection of whole-body human manipulation demonstrations, we develop a tracker-free capture system using only a consumer VR headset and a head-mounted stereo camera, without external cameras or motion trackers. We then perform three explicit alignment stages---embodiment, action-semantic, and temporal---to map captured motion into the robot action space, preserving fine-grained whole-body motion and allowing human and robot demonstrations to be jointly learned by standard VLA policies. Real-world experiments with different VLA backbones show that human demonstrations consistently improve policy learning across training paradigms, raising average success from 29\% to 56\% on GR00T N1.7 and from 32\% to 57\% on $\mathbf{\pi_{0.5}}$, while matching robot-only training with half the robot demonstrations. Ablations confirm that each alignment stage is necessary. These results highlight the potential of human demonstrations for scalable whole-body mobile manipulation with preserved fine-grained motion structure.
\end{abstract}

\setlength{\abovedisplayskip}{6pt plus 1pt}
\setlength{\belowdisplayskip}{10pt plus 1pt minus 1pt}
\setlength{\abovedisplayshortskip}{-10pt plus 1pt}
\setlength{\belowdisplayshortskip}{7pt plus 1pt minus 1pt}

\section{Introduction}
\label{sec:intro}

Mobile bimanual dexterous manipulation requires robots to simultaneously perform locomotion, torso adjustment, bimanual coordination, active vision, and finger-level dexterous interaction within a single trajectory. Unlike isolated manipulation or locomotion, these tasks require continuous coordination among multiple body components throughout execution. Training such systems therefore demands large-scale, high-quality whole-body action data. However, collecting robot demonstrations at this scale remains challenging as robot embodiments become increasingly complex. Existing mobile manipulation platforms, such as Human Support Robot~\cite{hsr} and Stretch~\cite{kemp2022designstretchcompactlightweight}, provide important foundations for embodied intelligence, while recent whole-body teleoperation systems extend data collection toward bimanual and humanoid manipulation through bilateral interfaces, kinematic-matched platforms, and wearable devices~\cite{fu2024mobilealohalearningbimanual,ben2025homiehumanoidlocomanipulationisomorphic,zhong2025humanoidexoscalablewholebodyhumanoid,zhong2025nuexowearableexoskeleton}. Nevertheless, these approaches require increasingly specialized hardware interfaces, making large-scale data collection for mobile bimanual dexterous robots expensive and difficult to scale.

Egocentric human demonstrations provide a scalable alternative for alleviating the robot data bottleneck. Large-scale egocentric datasets have revealed rich human activity priors~\cite{grauman2022ego4dworld3000hours,damen2020epickitchensdatasetcollectionchallenges,liu2024hoi4d4degocentricdataset}, while recent systems enable collecting robot-relevant human demonstrations through scalable interfaces~\cite{wang2024dexcapscalableportablemocap,chi2024universalmanipulationinterfaceinthewild,chen2024arcapcollectinghighqualityhuman,cheng2024opentelevisionteleoperationimmersiveactive}. Building on these advances, recent works demonstrate that human demonstrations can directly improve robot policy learning for dexterous manipulation~\cite{yuan2025motiontranshumanvrdata,zheng2026egoscalescalingdexterousmanipulation,tao2026dexwilddexteroushumaninteractions}, bimanual manipulation~\cite{bi2025hrdthumanmanipulationenhanced}, and vision-language-action models trained from egocentric data~\cite{yang2025egovlalearningvisionlanguageactionmodels}. However, directly transferring the full human motion space to robots remains challenging due to differences in morphology, control interfaces, and execution timescales between humans and robots.

To reduce this transfer difficulty, prior human-to-robot works often simplify or restructure human motion representations before policy learning. Instead of directly transferring the complete human motion space, these approaches introduce structured intermediate representations that reduce embodiment discrepancies and make robot learning more tractable. Recent mobile whole-body manipulation systems adopt different forms of such representation transformations, including simplified locomotion and hand representations~\cite{shi2026egohumanoidunlockinginthewildlocomanipulation}, phase-based decomposition of navigation and manipulation~\cite{zhu2025emmascalingmobilemanipulation}, and controller-assisted whole-body execution through task-level targets~\cite{xu2026hommilearningwholebodymobile,zhao2026halomilearninghumanoidlocomanipulation,wang2026bifrostumibridgingrobotfreedemonstrations,nai2026humanoidmanipulationinterfacehumanoid}. Other approaches avoid directly transferring human actions and instead leverage human videos for representation learning or foundation model pretraining~\cite{nair2022r3muniversalvisualrepresentation,ma2023vipuniversalvisualreward,ye2025latentactionpretrainingvideos,wei2026psi0openfoundationmodel}.

These representation transformations reduce the complexity of human--robot transfer, but they inherently introduce a trade-off between transferability and preserving human motion information. This trade-off is often acceptable for robots with limited action complexity and weak whole-body coupling, but becomes restrictive for mobile bimanual dexterous manipulation, where successful execution requires continuous coordination among locomotion, upper-body motion, and dexterous interaction. Specifically, existing approaches face several limitations. First, compressing continuous human motion into discrete or low-dimensional representations can remove fine-grained motion variations, preventing policies from fully exploiting the structure in human demonstrations. Second, decomposing long-horizon behaviors into independent phases can weaken dependencies between different stages of execution, making it difficult to jointly learn locomotion and upper-body manipulation within a unified policy. For example, separating navigation and manipulation limits the ability to perform tasks requiring tightly coupled arm--base coordination. Finally, in mobile bimanual manipulation, the motion of one body component often changes the feasible workspace and control constraints of others. Representing the base, torso, arms, and hands as independent control objectives ignores these interactions, limiting the ability to learn the whole-body coordination required for mobile dexterous manipulation.

This raises the central question of this work:
\textbf{Can human demonstrations be transferred to robots while resolving embodiment mismatch and preserving the fine-grained motion structure required for complex whole-body behaviors?}

We present \textbf{DexRoam} (Fig.~\ref{fig:teaser}), a complete system that answers this question by learning mobile bimanual dexterous manipulation from human demonstrations end to end, spanning portable egocentric capture, explicit whole-body human-to-robot alignment, and real-robot policy learning. The system is built around a single design principle: the continuous, coupled structure of whole-body motion is carried from the demonstrator's body to the robot's action chunks without discretized locomotion primitives, phase decomposition, or independent per-component control. Each stage is designed so that the continuous, coupled structure of whole-body motion survives from the demonstrator's body to the robot's action.

The first stage is a \textbf{tracker-free egocentric whole-body capture system}. It recovers whole-body human motion together with egocentric video using only a consumer VR headset and a head-mounted stereo camera, requiring no external cameras, environment markers or body-worn trackers. The camera stream is displayed inside the headset in real time, so the demonstrator's view during collection is exactly the observation the policy later sees. This keeps collection portable and scalable, while providing raw whole-body motion in human space as the input to the rest of the system.

The second stage is \textbf{explicit human-to-robot alignment}, which brings this motion into the robot's action space without compressing it. Three mismatches are resolved in turn. \emph{Embodiment mismatch}, from structural differences between humans and robots, is handled by root-centric whole-body retargeting. \emph{Action-semantic mismatch}, between human motion representations and robot control interfaces, is handled by robot-centric relative action construction. \emph{Temporal mismatch}, from the different execution timescales of humans and robots, is handled by task-progress temporal resampling. After alignment, human and robot demonstrations occupy the same observation--action space, so a standard VLA policy trains on both jointly, without human-specific action heads, and directly predicts complete whole-body action chunks.

We validate the full pipeline on a real mobile bimanual dexterous robot across five manipulation tasks, two VLA backbones, and varying robot-data budgets. Transferred human trajectories replay on the robot with centimeter-level error and no IK failures, and alignment substantially reduces the human--robot action distribution gap. Training on the aligned data improves policy performance consistently across tasks and backbones, and matches robot-only training with half the robot demonstrations. Ablations show that removing action-semantic or temporal alignment substantially degrades the policy, confirming that each stage is necessary for the pipeline to work. Our contributions are:

\begin{itemize}[leftmargin=*]
\item \textit{DexRoam}, a complete working system for learning mobile bimanual dexterous manipulation from human demonstrations, integrating portable egocentric capture, explicit whole-body alignment, and real-robot policy learning under a single continuous, coupled action interface.
\item \textit{A tracker-free egocentric whole-body capture system} that records full-body motion, hand articulation, and egocentric stereo video with only a consumer VR headset and a head-mounted camera, with in-headset streaming that keeps the demonstrator's view consistent with the policy observation.
\item \textit{An explicit three-stage alignment} resolving embodiment, action-semantic, and temporal mismatches while preserving continuous whole-body motion structure, so that human and robot data are interchangeable for a standard VLA policy.
\item \textit{Extensive real-world evaluation} across five tasks, two VLA backbones, and different robot-data budgets, covering trajectory executability, distribution alignment, policy improvement, data efficiency, and ablations.
\end{itemize}

\section{Problem Definition}
\label{sec:problem}
We study the transfer of human demonstrations to mobile bimanual dexterous robots. We define the robot whole-body action space as $\mathcal{A}$, spanning the mobile base, torso, dual arms, active head, and dexterous hands, and require that the transferred demonstrations satisfy a \emph{fine-grained whole-body action interface} with three properties:
\textbf{(1) Continuous action space.}  
All action dimensions remain continuous rather than being discretized.
\textbf{(2) Temporally continuous prediction.}  
The complete action sequence is continuously predicted by a single policy, without phase switching at deployment.
\textbf{(3) Coupled whole-body prediction.}  
Actions of all body components are jointly modeled in a unified policy output space, rather than being treated as independent control objectives during deployment.
Under this setting, we study how to effectively transfer human demonstrations to mobile bimanual dexterous robots while satisfying the above action interface constraints.

The training corpus consists of two sources: robot demonstrations $\mathcal{D}_{\mathrm{robot}}$ collected on the target embodiment, whose actions already lie in $\mathcal{A}$, and human demonstrations $\mathcal{D}_{\mathrm{human}}$, whose motions reside in human space. We therefore require a human-to-robot transformation $\mathcal{R}$ satisfying two conditions:

\begin{equation*}
\mathcal{R}(\mathcal{D}_{\mathrm{human}}) \subset \mathcal{S},
\qquad
p_{\mathcal{R}(\mathcal{D}_{\mathrm{human}})} \approx p_{\mathcal{D}_{\mathrm{robot}}},
\end{equation*}

where $\mathcal{S}$ is the observation--action space of the target embodiment, whose action component $\mathcal{A}$ satisfies the three properties above. 
The first condition makes both sources consumable by a single policy without human-specific action heads; 
the second ensures the transferred data do not pull the policy away from the action distribution it encounters at deployment. Only under both conditions can $\mathcal{R}(\mathcal{D}_{\mathrm{human}})$ and
$\mathcal{D}_{\mathrm{robot}}$ be trained on jointly under a single objective.
We define the policy as $\pi_\theta(\mathbf{A}_t \mid \mathbf{I}_t,\mathbf{P}_t,\ell)$, where the policy receives an egocentric visual observation $\mathbf{I}_t$, a proprioceptive state history $\mathbf{P}_t$, and a language instruction $\ell$, and outputs a continuous whole-body action chunk $\mathbf{A}_t\in\mathcal{A}$. 
After alignment, transformed human demonstrations join the training corpus $\mathcal{D}$ alongside robot demonstrations, enabling joint learning under the same policy and action interface.

\section{DexRoam}
\label{sec:method}

\begin{figure*}[t]
    \centering
    \includegraphics[width=\linewidth]{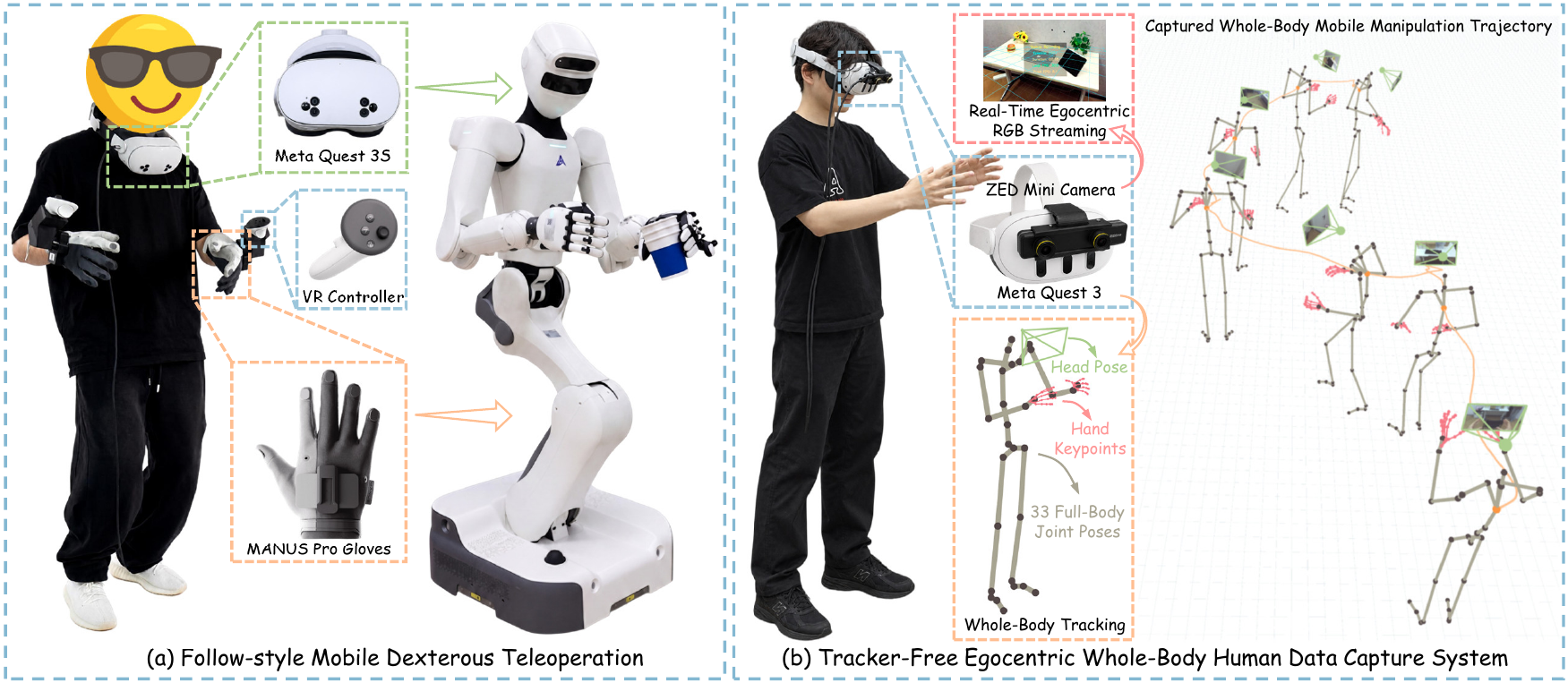}
    \vspace{-0.6cm}
    \caption{
\textbf{DexRoam data collection system.}
(a) Follow-style teleoperation lets one operator drive base, torso, arms, head, and dexterous hands simultaneously via a Meta Quest 3S, VR controllers, and MANUS Pro gloves, with gesture-based episode management. (b) Tracker-free egocentric capture uses a Meta Quest 3 and a head-mounted ZED Mini to record 6-DoF head pose, 33 body joints, 25 joints per hand, and timestamped egocentric stereo RGB, with no external cameras, marker or trackers. Real-time in-headset video streaming keeps the demonstrator's view identical to the policy observation.
}
    \label{fig:collection}
\vspace{-0.6cm}
\end{figure*}

In this section, we present DexRoam. Sec.~\ref{sec:collection} introduces the data collection system, Sec.~\ref{sec:alignment} presents the three-stage alignment addressing embodiment, action-semantic, and temporal mismatches, and Sec.~\ref{sec:policy} describes policy learning under the unified observation--action interface.

\subsection{Hardware Platform and Data Collection System}
\label{sec:collection}

\subsubsection{\textbf{Hardware platform}}
The Astribot--XHand platform comprises an omnidirectional mobile base, an actuated torso, two 7-DoF arms, a 2-DoF active head, and two 12-DoF XHand dexterous hands (Fig.~\ref{fig:collection}a, right), supporting coordinated whole-body mobile dexterous manipulation.

\subsubsection{\textbf{Robot demonstration collection}}
We build a whole-body teleoperation system (Fig.~\ref{fig:collection}a) for collecting robot demonstrations on the target mobile bimanual dexterous platform. The Meta Quest 3S captures upper-body poses, VR controllers provide wrist pose commands, and MANUS Pro gloves capture hand motions, with these inputs jointly mapped to the robot base, end-effectors, and XHand joints, enabling coupled whole-body mobile dexterous teleoperation. The system adopts a single-operator control paradigm, where the operator simultaneously controls robot locomotion and whole-body manipulation, eliminating the need for an additional operator dedicated to base navigation. Since wearing gloves makes conventional keyboard or button-based interaction inconvenient, we further design a gesture-based interface for episode state management, allowing the operator to start, stop, save, and discard episode directly through hand gestures.

\subsubsection{\textbf{Tracker-free egocentric whole-body data collection}}
Human demonstrations are captured with only a Meta Quest 3 headset and a head-mounted ZED Mini stereo camera (Fig.~\ref{fig:collection}b), requiring no external motion-capture cameras, no environment markers and no body-worn trackers. 
The Meta Quest 3 captures whole-body motion through WebXR~\cite{vuer}, including 6-DoF headset pose, 33 body-joint poses, and 25 hand landmarks per hand, which together form the whole-body human state $\mathbf{h}_t$. Meanwhile, the ZED Mini records egocentric stereo RGB images with timestamps, providing first-person visual observations for each demonstration. 
Fig.~\ref{fig:collection}b (right) shows a captured whole-body mobile manipulation trajectory, in which head pose, hand keypoints, and the full-body skeleton evolve continuously as the demonstrator moves through the scene. A gesture-based interface is also designed for episode management, allowing the operator to start, stop, save, and discard recordings directly through gestures.

The key design choice of the capture system is \emph{egocentric video streaming}: the ZED Mini stream is displayed inside the headset in real time (Fig.~\ref{fig:collection}b), so what the operator sees during collection is exactly the egocentric observation used later for policy training. This keeps visual observation, visual attention, and motor intent consistent throughout an episode. Such consistency is important for mobile dexterous manipulation, where the viewpoint continuously changes as the robot moves its base, body, and head.

\subsection{\textbf{Human-to-Robot Alignment}}
\label{sec:alignment}

DexRoam instantiates the human-to-robot transformation $\mathcal{R}$ through three explicit alignment steps(Fig.~\ref{fig:alignment}), which resolve embodiment, action-semantic, and temporal mismatches between human demonstrations and robot execution spaces.

\subsubsection{\textbf{Embodiment Alignment}}
\begin{figure*}[t]
    \centering
    \includegraphics[width=\linewidth]{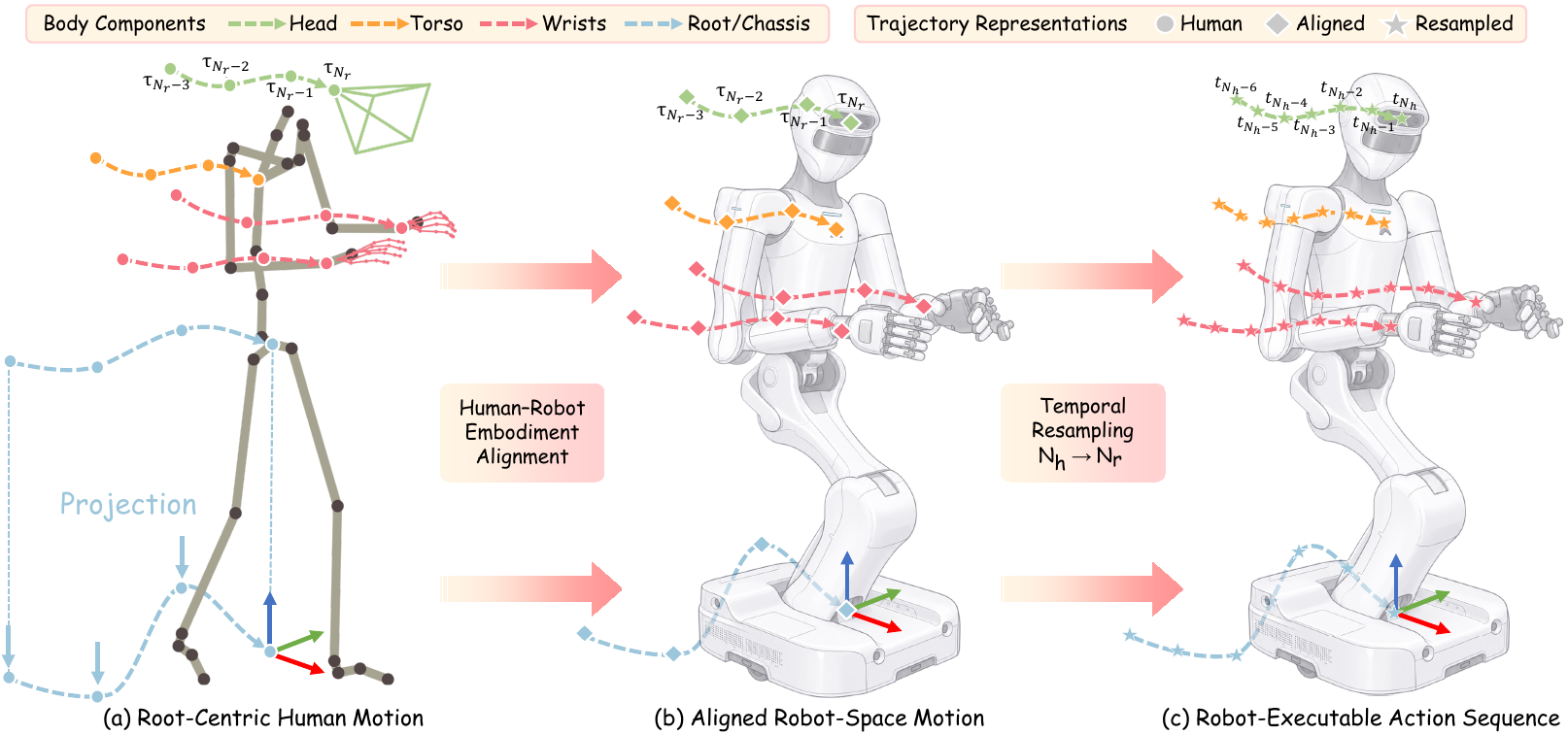}
    \vspace{-0.6cm}
    \caption{
            \textbf{Explicit human-to-robot alignment in DexRoam.}
            (a) Human motion is expressed in a pelvis-centric root frame projected onto the ground plane, yielding root/chassis, torso, wrist, and head trajectories free of global scene coordinates. (b) Embodiment alignment maps each component to its robot counterpart, and action-semantic alignment re-expresses the results as robot-centric relative transformations. (c) Temporal alignment resamples $N_h$ human steps into $N_r$ waypoints evenly spaced in task progress, matching the execution timescale of robot demonstrations.
            }
    \label{fig:alignment}
\vspace{-0.6cm}
\end{figure*}

Humans and robots have different kinematic trees, so neither joint angles nor global poses can be copied directly into an executable action. To reduce dependence on global coordinate frames and preserve body-relative motion structure, we introduce a time-varying pelvis-centric root frame $\mathcal{F}_{\mathrm H}$ to align human motion with the robot embodiment. The origin of $\mathcal{F}_{\mathrm H}$ is the orthogonal projection of the pelvis onto the ground plane (\emph{Projection} in Fig.~\ref{fig:alignment}a); its forward axis is the horizontal projection of the pelvis's local forward direction, and its vertical axis points opposite to gravity. Let $\mathbf{z}_{\mathrm H}=-\widehat{{}^{\mathrm W}\mathbf{g}}$ denote the upward axis of the
human root frame, where ${}^{\mathrm W}\mathbf{g}$ is the gravity vector expressed in
the world frame. The ground-plane projection operator is defined as
$\mathbf{\Pi}_{\mathrm G}=\mathbf{I}-\mathbf{z}_{\mathrm H}\mathbf{z}_{\mathrm H}^\top$. Assuming that the world origin lies
on the ground plane, the pose of $\mathcal{F}_{\mathrm H}$ is defined as:

\begin{equation*}
\setlength{\arraycolsep}{3pt}
\begin{gathered}
\mathbf{x}_t
=
\widehat{\mathbf{\Pi}_{\mathrm G}\,{}^{\mathrm W}\mathbf{f}_{\mathrm{pelvis},t}},
\qquad
\mathbf{y}_t
=
\mathbf{z}_{\mathrm H}\times\mathbf{x}_t,\\[0.25em]
{}^{\mathrm W}\mathbf{T}_{\mathrm H,t}
=
\left[
\begin{array}{@{}cccc@{}}
\mathbf{x}_t & \mathbf{y}_t & \mathbf{z}_{\mathrm H} &
\mathbf{\Pi}_{\mathrm G}\,{}^{\mathrm W}\mathbf{p}_{\mathrm{pelvis},t}\\
0 & 0 & 0 & 1
\end{array}
\right]
\end{gathered}
\end{equation*}
Aligned with the robot base motion plane, $\mathcal{F}_{\mathrm H}$ establishes a human-centric reference frame shared by locomotion and manipulation. It serves two purposes: transferring the human locomotion trajectory to the robot chassis motion and representing upper-body movements as body-relative trajectories, reducing dependence on global scene coordinates. Specifically, torso motion, head orientation, and wrist trajectories are expressed in $\mathcal{F}_{\mathrm H}$ through

\begin{equation*}
{}^{\mathrm H}\mathbf{T}_{J,t}=({}^{\mathrm W}\mathbf{T}_{\mathrm H,t})^{-1}\,{}^{\mathrm W}\mathbf{T}_{J,t},
\end{equation*}
where $J \in \{\mathrm{chest},\mathrm{head},\mathrm{l{-}wrist},\mathrm{r{-}wrist}\}$. The resulting relative trajectories are then converted into robot commands: the human root trajectory is mapped to the chassis motion, the chest pose is used as the torso target, and the left/right wrist poses are transformed into arm end-effector targets. For the head, although its pose is represented in $\mathcal{F}_{\mathrm H}$, only the yaw and pitch components are extracted as the 2-DoF head command due to the robot's mechanical constraints. Fig.~\ref{fig:alignment}b illustrates the resulting per-component trajectories in robot space. For the dexterous hands, we adopt dex-retargeting~\cite{qin2023anyteleop} to retarget human hand poses into the 12-DoF XHand commands, bridging the morphology gap between human and robot hands.

\subsubsection{\textbf{Action-Semantic Alignment}}
Retargeting yields poses, not actions: the goal is not merely to reach the
robot's pose space, but to construct targets carrying the same semantics as
robot control. Absolute poses fail this test, as they still encode
demonstrator-specific configuration such as height, shoulder width, arm length,
and initial posture. We therefore express continuous motion as relative
transformations in each body component's local frame, which specify how a
component moves from where it currently is rather than where it should be in
absolute coordinates, and are thus independent of the configuration in which
the demonstration was recorded. For upper-body components, poses are already represented in the human root frame
$\mathcal{F}_{\mathrm H}$, we compute the relative motion for
$J\in\{\mathrm{chest},\mathrm{l{-}wrist},\mathrm{r{-}wrist}\}$ as

\begin{equation*}
\Delta\mathbf{T}^{J}_{t}
=({}^{\mathrm H}\mathbf{T}_{J,t})^{-1}{}^{\mathrm H}\mathbf{T}_{J,t+1}
\end{equation*}
The resulting local transformations describe how each component evolves from
its current state and preserve transferable motion patterns such as reaching,
lifting, and retracting.

For locomotion, the root frame encodes the human movement. We compute the relative transformation between consecutive root frames and represent it as planar motion:
\vspace{-0.05cm}
\begin{equation*}
\Delta\mathbf{T}^{\mathrm H}_{t}
=({}^{\mathrm W}\mathbf{T}_{\mathrm H,t})^{-1}{}^{\mathrm W}\mathbf{T}_{\mathrm H,t+1},
\Delta\boldsymbol{\xi}^{\mathrm H}_{t}
=[\Delta x_t,\Delta y_t,\Delta\psi_t]^{\top}
\end{equation*}
Since $\mathcal{F}_{\mathrm H}$ is constructed on the ground plane with a
gravity-aligned vertical axis, its relative motion is inherently planar.
Compared with absolute trajectories, this representation removes dependence
on scene coordinates and initial positions while providing explicit
locomotion semantics, enabling continuous coupling between base movement and
whole-body manipulation.
The 2-DoF head command and the 12-DoF hand command already lie in robot joint
space and are used directly.

\subsubsection{\textbf{Temporal Alignment}}
\label{sec:temporal}

We align demonstrations by task progress rather than wall-clock time. Humans complete the same task faster and in fewer steps than the robot, so the two advance task progress at different rates per policy step.
We resample each retargeted human trajectory into $\bar{N}_r$ evenly spaced waypoints in normalized task progress $s\in[0,1]$, where $\bar{N}_r$ is the mean robot episode length for that task:
\begin{equation*}
\tilde{T}_k=\operatorname{Interp}(T_{1:N_h},s_k),\;
s_k=\frac{k-1}{\bar N_r-1},\;1\leq k\leq\bar N_r
\end{equation*}
with $N_h$ the original human trajectory length. After resampling, the two sources advance the same fraction of task progress per policy step, as illustrated in Fig.~\ref{fig:alignment}c. The resampled waypoints preserve the original spatial trajectory while matching the temporal distribution of robot executions. Positions are interpolated linearly and rotations by SLERP, while images and discrete signals use nearest-neighbor matching; base yaw is unwrapped before interpolation to avoid discontinuities around $\pm\pi$. Resampling only adjusts the temporal discretization of the trajectories and does not modify the robot control frequency or action chunk size.

After alignment, human and robot demonstrations become interchangeable under the same policy interface. The resulting action representation satisfies the three properties defined in Sec.~\ref{sec:problem}. First, all actions remain continuous without discretization. Second, the complete whole-body trajectory is predicted by a single policy without phase-based execution. Third, all body components are jointly represented in one unified action space. This unified interface enables human demonstrations to retain their fine-grained motion structure while becoming executable in the robot action space.


\subsubsection{Policy Learning}
\label{sec:policy}

Because aligned human data and robot data share the same observation--action space, they can be directly consumed by a standard VLA policy without human-specific action heads or input branches. We instantiate $\pi_\theta(A_t|I_t,P_t,\ell)$ with two VLA backbones, GR00T N1.7~\cite{gr00tn1_2025} and $\pi_0.5$~\cite{intelligence2025pi05visionlanguageactionmodelopenworld}. Each policy takes egocentric RGB observations, proprioceptive state history, and language instructions as inputs, and predicts whole-body action chunks.
We design three strategies for incorporating aligned human demonstrations into policy learning:
\textbf{Human--Robot Cotrain}, jointly training on aligned human and robot demonstrations;
\textbf{Human Pretrain + Robot FT}, pretraining on aligned human demonstrations and finetuning on robot demonstrations only;
and \textbf{Human Pretrain + Cotrain FT}, pretraining on aligned human demonstrations and finetuning on the human--robot mixture.
All strategies use the same observation--action interface and downstream finetuning budget.
\section{Experiments}
\label{sec:exp}

\begin{figure*}[t]
    \centering
    \includegraphics[width=0.95\linewidth]{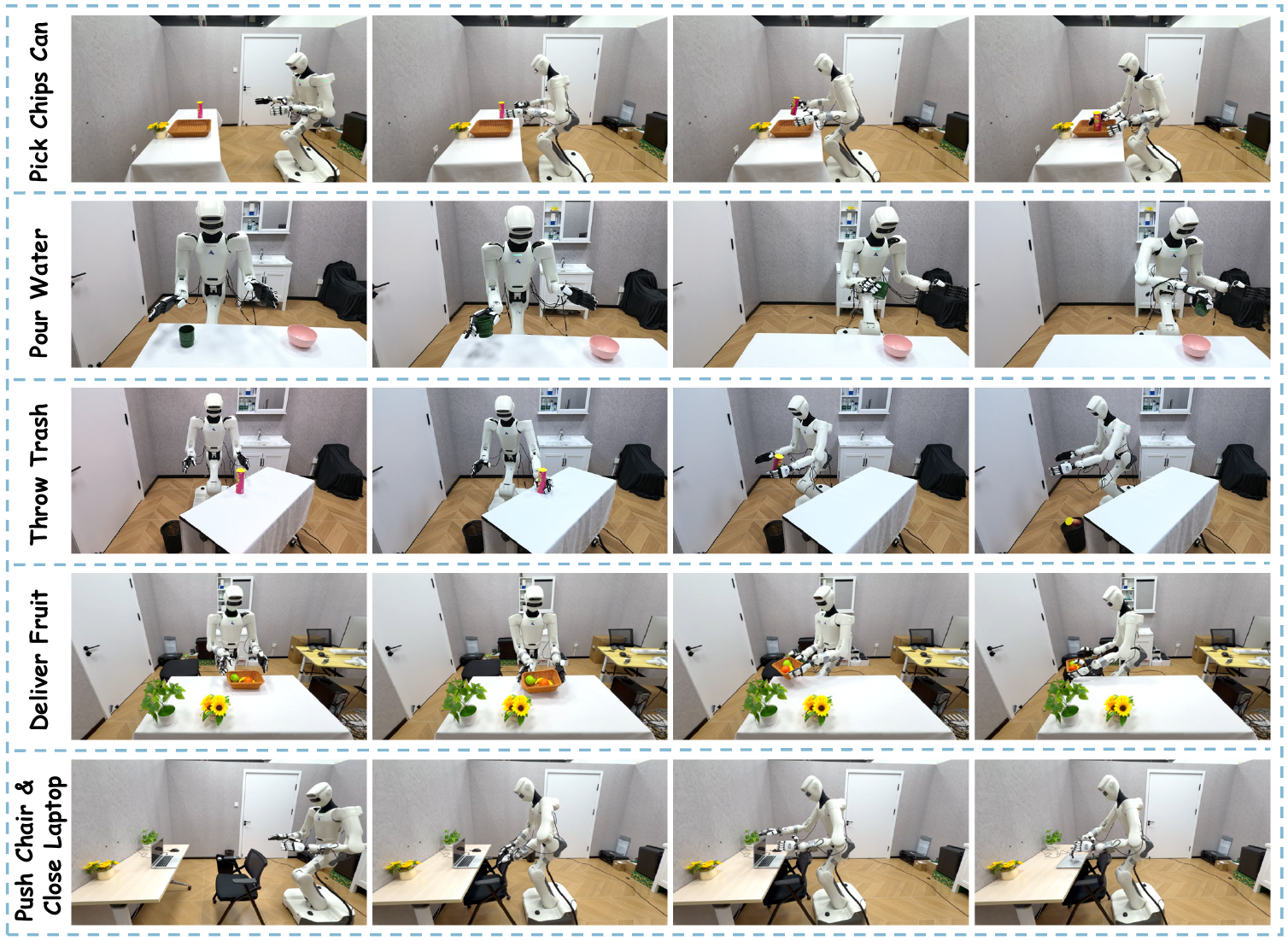}
    \vspace{-0.3cm}
    \caption{
            \textbf{Real-world mobile bimanual dexterous manipulation tasks.}
Rollouts of the five evaluation tasks, one per row: Pick Chips Can, Pour Water, Throw Trash, Deliver Fruit, and Push Chair \& Close Laptop. They span fine-grained dexterity, base–manipulation coupling, and long-horizon whole-body interaction.
            }
    \label{fig:tasks}
\vspace{-0.6cm}
\end{figure*}

\subsection{Experimental Setup}
\label{sec:exp_setup}
\subsubsection{\textbf{Tasks}}
We evaluate DexRoam on five tasks covering complementary mobile manipulation capabilities (Fig.~\ref{fig:tasks}).
\textbf{Pick Chips Can} requires the robot to navigate to a table, grasp a chips can, and place it in a tray, emphasizing precise reaching and finger-level grasping.
\textbf{Pour Water} requires grasping a cup and pouring into a target container, testing continuous wrist--hand coordination.
\textbf{Throw Trash} requires grasping an object, moving to a bin, and releasing it accurately; the object must remain stable during locomotion, while successful disposal depends on precise coordination between base positioning and manipulation.
\textbf{Deliver Fruit} requires carrying a basket bimanually to a target workstation, stressing bimanual transport and coordinated turning during locomotion.
\textbf{Push Chair \& Close Laptop} chains long-range navigation, contact-rich chair pushing, and laptop closing, testing long-horizon execution together with continuous base--upper-body coordination and fine manipulation.


\begin{figure*}[t]
    \centering
    \includegraphics[width=1.0\textwidth]{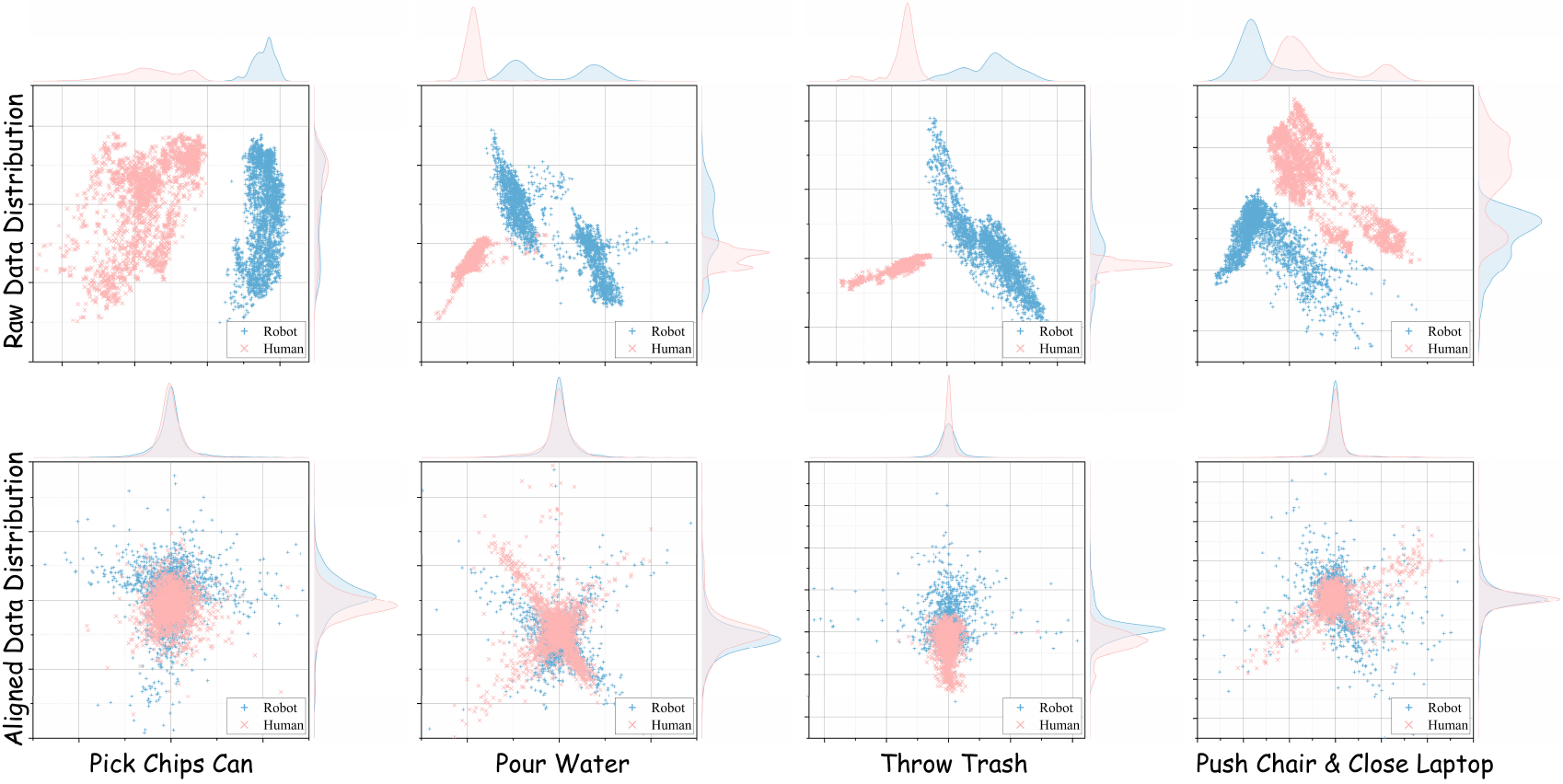}
    \vspace{-0.8cm}
    \caption{
            \textbf{Human--robot action distribution before and after alignment.}
            PCA visualization of whole-body actions on representative tasks. Top: retargeted absolute human actions and robot actions before action-semantic and temporal alignment, which occupy clearly separated regions. Bottom: fully aligned robot-centric relative actions, where the two distributions largely overlap, with marginals shown along each axis.
            }
    \label{fig:action_distribution}
    \vspace{-0.4cm}
\end{figure*}

\begin{figure}[t]
    \centering
    \includegraphics[width=1.0\linewidth]{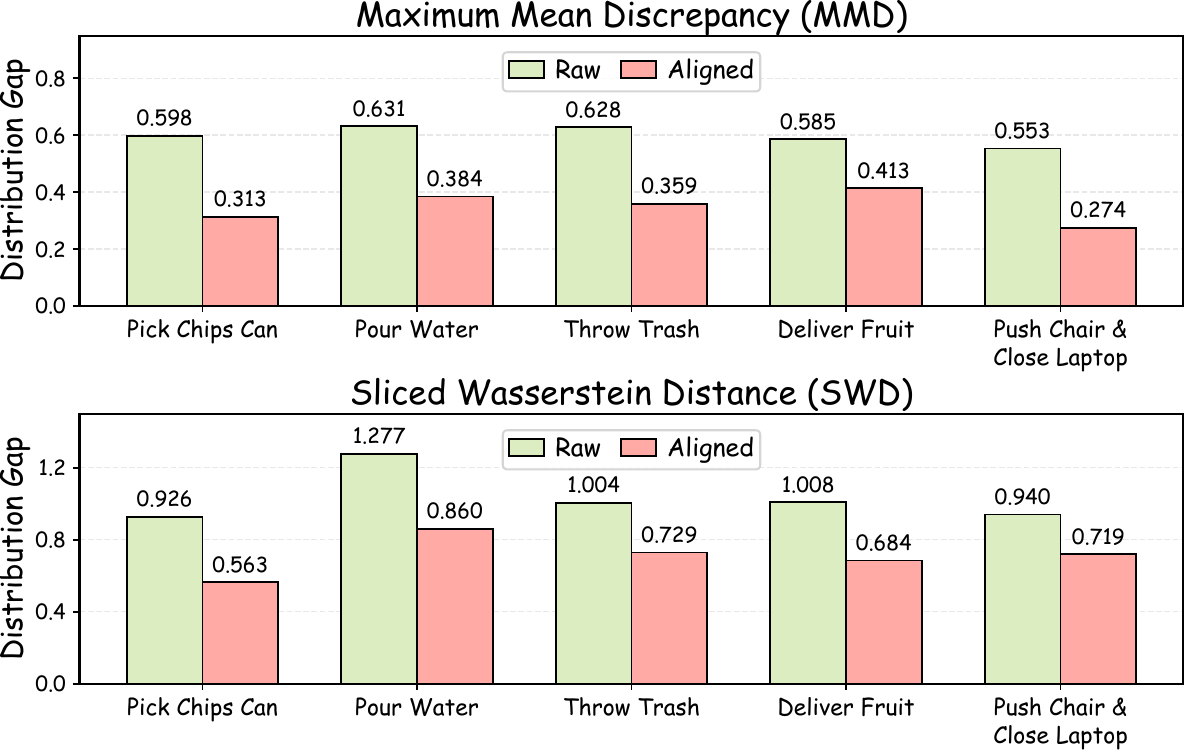}
    \vspace{-0.6cm}
    \caption{
            \textbf{Quantitative human--robot action distribution alignment.}
            MMD and SWD between human-derived and robot whole-body actions on all five tasks, lower values indicate closer distributions. Alignment reduces both metrics on every task, by 41.7\% in MMD and 31.1\% in SWD on average.
            }
            \vspace{-0.6cm}
    \label{fig:mmd_swd_bar}
\end{figure}

\subsubsection{\textbf{Evaluation protocol}}
We evaluate DexRoam at three complementary levels: trajectory executability, human--robot action distribution alignment, and policy performance.

\noindent\textbf{Trajectory replay.}
We directly replay transferred human trajectories on the real robot and report trajectory completion, IK failure rate, and positional and rotational replay errors for the torso, arms, and head.

\noindent\textbf{Action distribution alignment.}
We further examine whether the alignment process brings human-derived actions closer to real robot actions.
For each task, we compare the human and robot whole-body action distributions at the retargeted-absolute stage and after full alignment.
We use Maximum Mean Discrepancy (MMD) and Sliced Wasserstein Distance (SWD) for quantitative comparison and PCA for qualitative visualization.
Detailed normalization, metric computation, and visualization protocols are provided in the Appendix.

\noindent\textbf{Policy evaluation.}
We evaluate all four training settings described in Sec.~\ref{sec:policy} on all five tasks, with 20 real-world trials per task.
Success rate (SR) captures whether the final task goal is achieved, but provides limited information about partial progress, particularly for mobile manipulation tasks.
We therefore additionally use task completion score (TCS), which decomposes each task into ordered stages and measures the fraction of stages completed. Detailed TCS criteria are provided in the Appendix. We report both SR and TCS for downstream policy evaluation.
Based on these evaluation metrics, we investigate when, how, and why aligned human demonstrations benefit mobile bimanual dexterous manipulation. Specifically, we study four questions:

\begin{itemize}
    \item \textbf{Q1: Transfer Quality.}
    How effectively does human-to-robot transfer convert human motion into executable and distribution-aligned robot training data?

    \item \textbf{Q2: Policy Learning.}
    Does aligned human supervision improve policy learning, and which training paradigm leverages it most effectively?

    \item \textbf{Q3: Data Efficiency.}
    How does aligned human data improve robot learning with limited robot demonstrations?

    \item \textbf{Q4: Alignment Analysis.}
    How do different alignment choices affect fine-grained human-to-robot transfer?

\end{itemize}

\begin{table}[t]
\centering
\caption{\textbf{Real-robot replay of transferred human trajectories.}
Position and rotation errors of retargeted human motions across different body components, measured in millimeters (mm) and degrees ($\degree$).}
\vspace{-0.2cm}
\label{tab:replay}
\setlength{\tabcolsep}{0pt}
\renewcommand{\arraystretch}{1}
\begin{tabular*}{\columnwidth}{@{}l@{\hspace{5pt}}c@{\extracolsep{\fill}}cccc@{}}
\toprule
\multicolumn{1}{c}{Component} & Metric & Pick Chips Can & Pour Water & Throw Trash & Average \\
\midrule
\multirow{2}{*}{Torso} & Position & 13.49 & 11.82 & 12.31 & \textbf{12.54} \\
& Rotation & 2.32 & 1.64 & 1.57 & \textbf{1.84} \\
\midrule
\multirow{2}{*}{Left Arm} & Position & 3.75 & 4.27 & 4.02 & \textbf{4.01} \\
& Rotation & 0.51 & 0.58 & 0.57 & \textbf{0.55} \\
\midrule
\multirow{2}{*}{Right Arm} & Position & 5.30 & 5.93 & 7.67 & \textbf{6.30} \\
& Rotation & 0.70 & 0.84 & 1.22 & \textbf{0.92} \\
\midrule
Head & Rotation & 1.22 & 1.36 & 1.53 & \textbf{1.37} \\
\midrule
Completed & & 5/5 & 5/5 & 5/5 & \textbf{15/15} \\
\bottomrule
\end{tabular*}
\vspace{-0.4cm}
\end{table}

\begin{table}[t]
\centering
\caption{\textbf{Tracker-free vs.\ tracker-based capture.} Retargeting accuracy averaged over 15 trajectories across three tasks. Position and rotation errors are reported in millimeters and degrees, respectively.}
\vspace{-0.2cm}
\label{tab:tracker}
\setlength{\tabcolsep}{1.5pt}
\renewcommand{\arraystretch}{1.15}
\begin{tabular*}{\columnwidth}{@{}c@{\extracolsep{\fill}}*{7}{c}@{}}
\toprule
\multirow[c]{2}{*}{\shortstack[c]{Human Motion\\Capture}} & \multicolumn{2}{c}{Torso} & \multicolumn{2}{c}{Left Arm} &
\multicolumn{2}{c}{Right Arm} & Head \\
\noalign{\vskip-\aboverulesep}
\cmidrule(lr){2-3}\cmidrule(lr){4-5}\cmidrule(lr){6-7}\cmidrule(l){8-8}
\noalign{\vskip-\belowrulesep}
& Pos. & Rot. & Pos. & Rot. & Pos. & Rot. & Rot. \\
\midrule
Tracker-free (Ours) & \textbf{12.54} & \textbf{1.84}
& \textbf{4.01} & \textbf{0.55} & 6.30 & 0.92 & 1.37 \\
PICO + Motion Trackers & 13.16 & 2.06 & 4.12 & 0.57
& \textbf{5.63} & \textbf{0.81} & \textbf{1.23} \\
\bottomrule
\end{tabular*}
\vspace{-0.4cm}
\end{table}

\subsection{Q1: Transfer Quality}
\label{sec:exp_q1}

\subsubsection{\textbf{Action distribution alignment}}
We examine whether the transferred human supervision is compatible with the action distribution of the real robot.
Across all five tasks, alignment consistently reduces both distribution metrics, decreasing MMD by \textbf{41.7\%} and SWD by \textbf{31.1\%} on average (Fig.~\ref{fig:mmd_swd_bar}).
The PCA visualization in Fig.~\ref{fig:action_distribution} shows the same qualitative trend, with substantially greater overlap between human-derived and robot actions after alignment.
Together with the replay results, these findings show that DexRoam produces human supervision that is not only executable on the target embodiment, but also better matched to real robot actions.

\subsubsection{\textbf{Real-robot executability}}
All 15 transferred human trajectories replay successfully on the real robot with no IK failures (Tab.~\ref{tab:replay}).
Across the three tasks, positional replay errors remain at approximately the centimeter scale for the torso and millimeter scale for both arms, while rotational errors remain within a few degrees across all evaluated body components.
Despite the different whole-body motion patterns involved, the transferred trajectories can therefore be consistently executed by the target robot, supporting their use as robot supervision.

\subsubsection{\textbf{Tracker-free capture fidelity}}
Since tracker-free human motion capture is a key component of our data collection system, we further examine whether this lightweight design compromises the quality of the transferred trajectories.
For comparison, we construct a tracker-based counterpart using a PICO headset with five Motion Trackers and evaluate it under the same replay protocol.
Both systems successfully replay all trajectories and achieve comparable replay errors across body components, with no systematic performance gap (Tab.~\ref{tab:tracker}).
This suggests that our tracker-free system substantially reduces sensing cost and wearable burden while preserving the quality required for subsequent human-to-robot transfer.

\begin{figure*}[t]
    \centering
    \includegraphics[width=\linewidth]{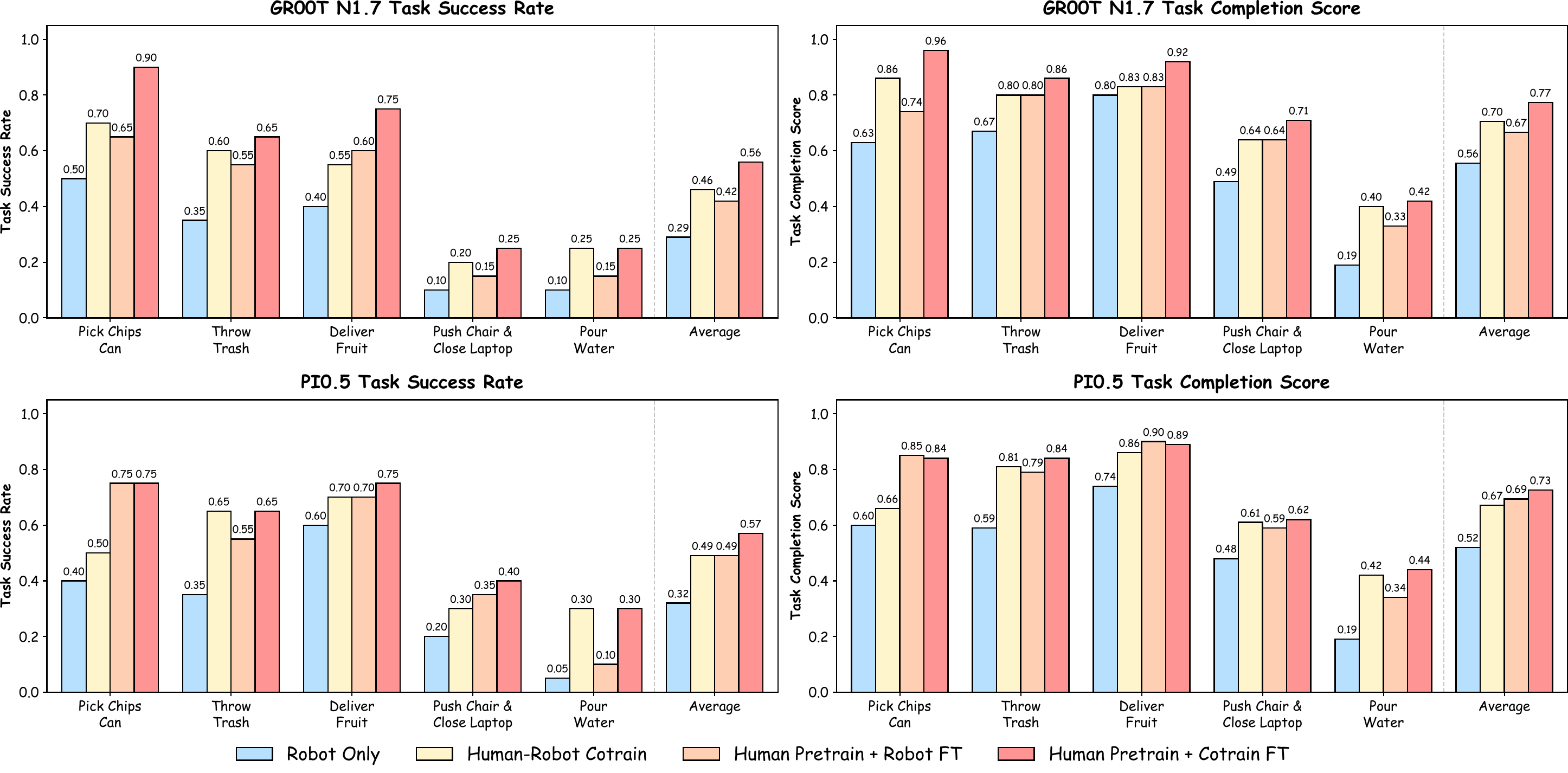}
    \vspace{-0.6cm}
    \caption{
            \textbf{Main policy learning results.}
            Success rate (SR, left) and task completion score (TCS, right) of Robot-only and three human-assisted training strategies on five real-world tasks, for GR00T N1.7 (top) and $\pi_{0.5}$ (bottom). All three human-assisted strategies improve over Robot-only on every task and both backbones, with Human Pretrain + Cotrain FT best or tied-best on every task–backbone combination.
            }
    \label{fig:main}
\vspace{-0.4cm}
\end{figure*}

\subsection{Q2: Policy Learning}
\label{sec:exp_q2}

Relative to Robot-only, all three human-assisted training paradigms improve both SR and TCS across both VLA backbones and all five tasks (Fig.~\ref{fig:main}).
In terms of average SR, the three paradigms provide gains of \textbf{+17, +13, and +27 percentage points} on GR00T N1.7, and \textbf{+17, +17, and +25 percentage points} on $\pi_{0.5}$.
The same positive trend holds across individual tasks, indicating that the benefit of aligned human supervision is not tied to a particular training paradigm, policy backbone, or manipulation setting.
This remains true on Push Chair \& Close Laptop, the longest-horizon and most tightly coupled task, where the best human-assisted setting improves success by \textbf{15 percentage points} on GR00T N1.7 and \textbf{20 percentage points} on $\pi_{0.5}$.
Together, these results show that aligned human supervision provides a broadly useful learning signal for mobile bimanual dexterous manipulation.

The magnitude of the gain, however, depends on how the human supervision is used.
Human Pretrain + Cotrain FT achieves the highest average performance on both backbones and is best or tied-best on every task--backbone combination.
This suggests that using human data both to initialize task-relevant behavior and to retain human supervision during subsequent human--robot cotraining makes the most effective use of the transferred demonstrations.

Qualitatively, we also observe that human supervision can introduce useful motion variations absent from the robot demonstrations.
On Pick Chips Can, the Robot-only policy predominantly approaches the can from the front and often knocks it over when its orientation is perturbed.
With human supervision, the policy can instead adopt a side approach when appropriate, selecting different grasp motions according to the object orientation.
Since this side-grasp pattern is absent from the robot demonstrations but present in the human data, the behavior provides a concrete example of how human supervision can expand the motion patterns available to the learned policy.

\subsection{Q3: Data Efficiency}
\label{sec:exp_q3}

\begin{figure}[t]
    \centering
    \includegraphics[width=1.0\linewidth]{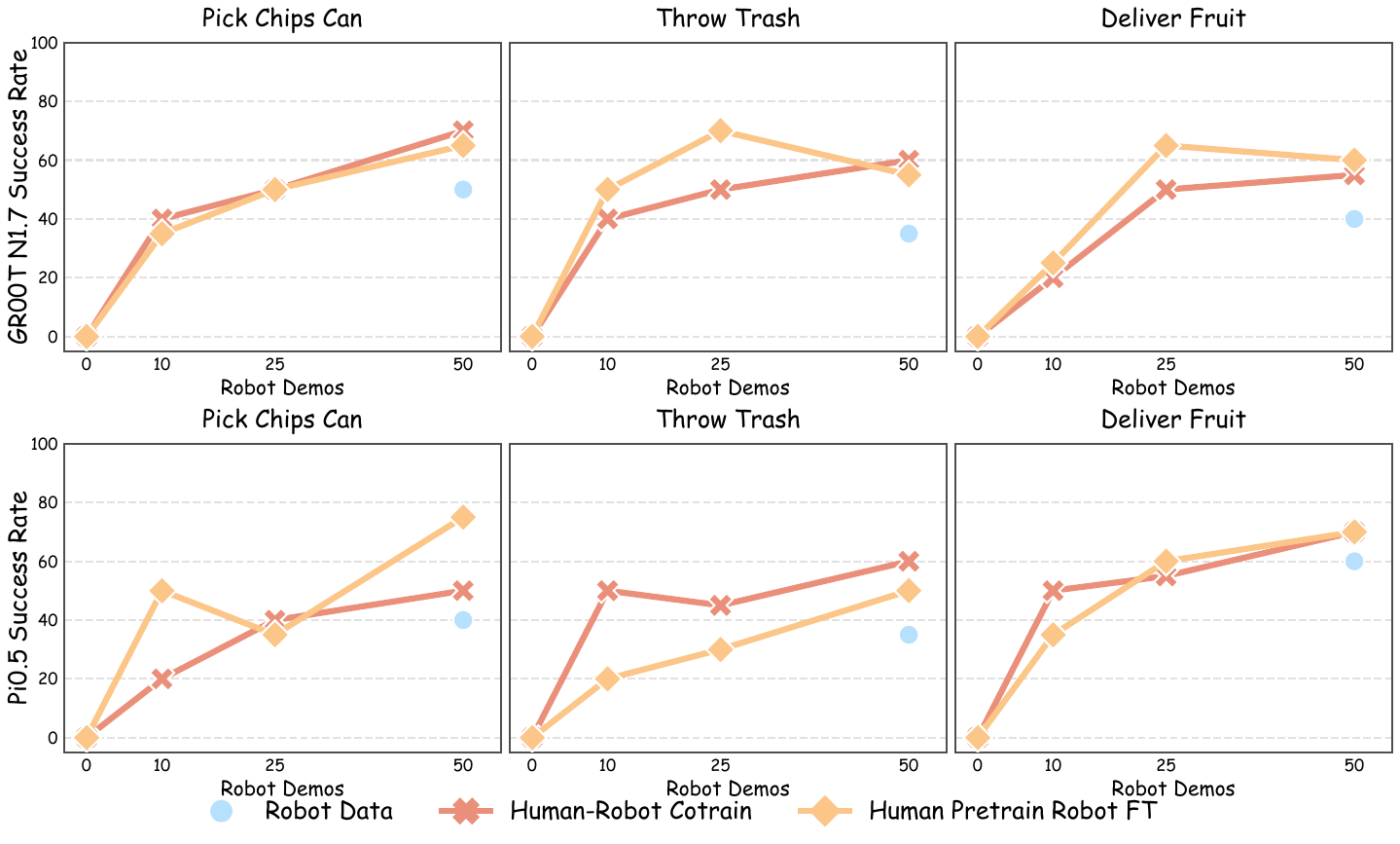}
    \vspace{-0.6cm}
    \caption{\textbf{Robot-data efficiency with human supervision.} Success rate under robot demonstration budgets of $\{0, 10, 25, 50\}$ with 50 human demonstrations fixed per task. With aligned human data, success rises sharply once only a few robot demonstrations are available, and 25 already approach or match the 50-demo Robot-only baseline.}
\label{fig:data_efficiency}
\vspace{-0.6cm}
\end{figure}

We fix 50 human demonstrations per task, vary the number of robot demonstrations over $\{0, 10, 25, 50\}$, and evaluate different human supervision training strategies, with Robot-only using 50 robot demonstrations as the reference (Fig.~\ref{fig:data_efficiency}).

\subsubsection{\textbf{Human data reduces robot data requirements}}
With only 25 robot demonstrations, human-assisted training on GR00T N1.7 already matches or exceeds the 50-demo Robot-only baseline on all three tasks, effectively \textbf{reducing the required robot demonstrations by 50\%}.
On $\pi_{0.5}$, the same 25-demo setting remains within 5 percentage points of the 50-demo Robot-only baseline.
These results show that aligned human supervision substantially improves robot-data efficiency, enabling comparable policy performance with fewer robot demonstrations.

\subsubsection{\textbf{Few-shot robot data activates human priors}}
With no robot demonstrations, neither human-assisted strategy achieves any task success.
However, performance rises sharply once only 10 robot demonstrations are introduced.
This suggests that aligned human data provides useful task and motion priors that can be activated with only a small amount of robot-specific supervision.
At the same time, the failure of the human-only setting shows that human data amplifies rather than replaces robot data: robot demonstrations remain necessary to ground the policy in the target embodiment and execution domain.

The two training paradigms also exhibit different scaling behavior.
Cotrain generally improves more steadily as robot data increases, whereas Human Pretrain + Robot FT is less stable.
We hypothesize that finetuning exclusively on a limited robot dataset can partially overfit to the small robot distribution and weaken the diversity introduced during human pretraining, while cotraining continuously retains human supervision throughout optimization.

\subsection{Q4: Alignment Analysis}
\label{sec:exp_q4}

\begin{table}[t]
\centering
\caption{\textbf{Ablation of alignment stages.} GR00T N1.7 success rate.}
\vspace{-0.2cm}
\label{tab:ablation}
\setlength{\tabcolsep}{4pt}
\begin{tabular}{@{}llccc@{}}
\toprule
\multirow{2}{*}{Task} & \multirow{2}{*}{Strategy} & w/o Temporal & Absolute & \textbf{DexRoam} \\
 & & Resampling & Action & \textbf{(Ours)} \\
\midrule
\multirow{3}{*}{\begin{tabular}[c]{@{}l@{}}Pick\\ Chips Can\end{tabular}}
& Cotrain & 0.65 & 0.00 & \textbf{0.70} \\
& Pretrain + FT & 0.30 & 0.00 & \textbf{0.65} \\
& Pretrain + Cotrain FT & 0.20 & 0.00 & \textbf{0.90} \\
\midrule
\multirow{3}{*}{\begin{tabular}[c]{@{}l@{}}Throw\\ Trash\end{tabular}}
& Cotrain & 0.40 & 0.00 & \textbf{0.65} \\
& Pretrain + FT & 0.50 & 0.00 & \textbf{0.55} \\
& Pretrain + Cotrain FT & 0.60 & 0.00 & \textbf{0.65} \\
\bottomrule
\end{tabular}
\vspace{-0.4cm}
\end{table}

We ablate the two alignment stages that can be removed in isolation, on Pick Chips Can and Throw Trash (Tab.~\ref{tab:ablation}).

\subsubsection{Absolute actions fail without semantic alignment}
Replacing DexRoam's robot-centric relative action representation with absolute retargeted
poses causes the policy to fail completely: success is 0\% on both tasks and under all
three training strategies (Tab.~\ref{tab:ablation}). We attribute this failure to the fact
that absolute poses map human and robot demonstrations into a shared coordinate
representation while still retaining a substantial amount of human-specific information.
During training, these factors pull the supervision signal away from the robot action
distribution the policy will face at deployment, so the policy learns targets entangled
with human-specific configurations rather than a stable robot control semantics.

This distribution shift is further amplified over closed-loop execution. At the start of
an episode, the robot's initial observation $\mathbf{o}_0=(\mathbf{I}_0,\mathbf{P}_0)$ lies within the
distribution of the robot data, so the first predicted chunk $\mathbf{a}_{0:H-1}$ still exhibits a
plausible execution trend. However, because the training distribution has been pulled
toward that of human-specific absolute actions, the model biases its predicted targets
toward the demonstrator's absolute configurations and incurs a small error. Executing
this error drives the next observation $\mathbf{o}_H=(\mathbf{I}_H,\mathbf{P}_H)$ off the training distribution,
which in turn makes the following chunk $\mathbf{a}_{H:2H-1}$ more erroneous. Repeated over
an episode, this forms a self-amplifying feedback loop that eventually drives the robot
into a state from which it cannot recover, and the task fails.

Among the three strategies, Human Pretrain + Robot FT exhibits this behavior to a
somewhat lesser degree under the absolute-action setting, since finetuning on robot data
pulls the learned distribution partly back toward the robot's own. The finetuning budget,
however, is not sufficient to recover it fully, and the resulting policy still completes
no trials.

\subsubsection{Temporal alignment improves policy learning}

Removing task-progress temporal resampling reduces the average success rate from 68.3\% to 44.2\%, a drop of \textbf{24.1 percentage points} (Tab.~\ref{tab:ablation}).
This degradation is caused by the different execution timescales of human and robot demonstrations: without resampling, the same task progress can correspond to different future action chunks, introducing temporally inconsistent supervision.
Task-progress temporal resampling aligns human and robot trajectories at the same execution phase, removing this temporal mismatch.

\subsection{Generalization Under Distribution Shift}
\label{sec:exp_q5}

\begin{figure}[t]
    \centering
    \includegraphics[width=1.0\linewidth]{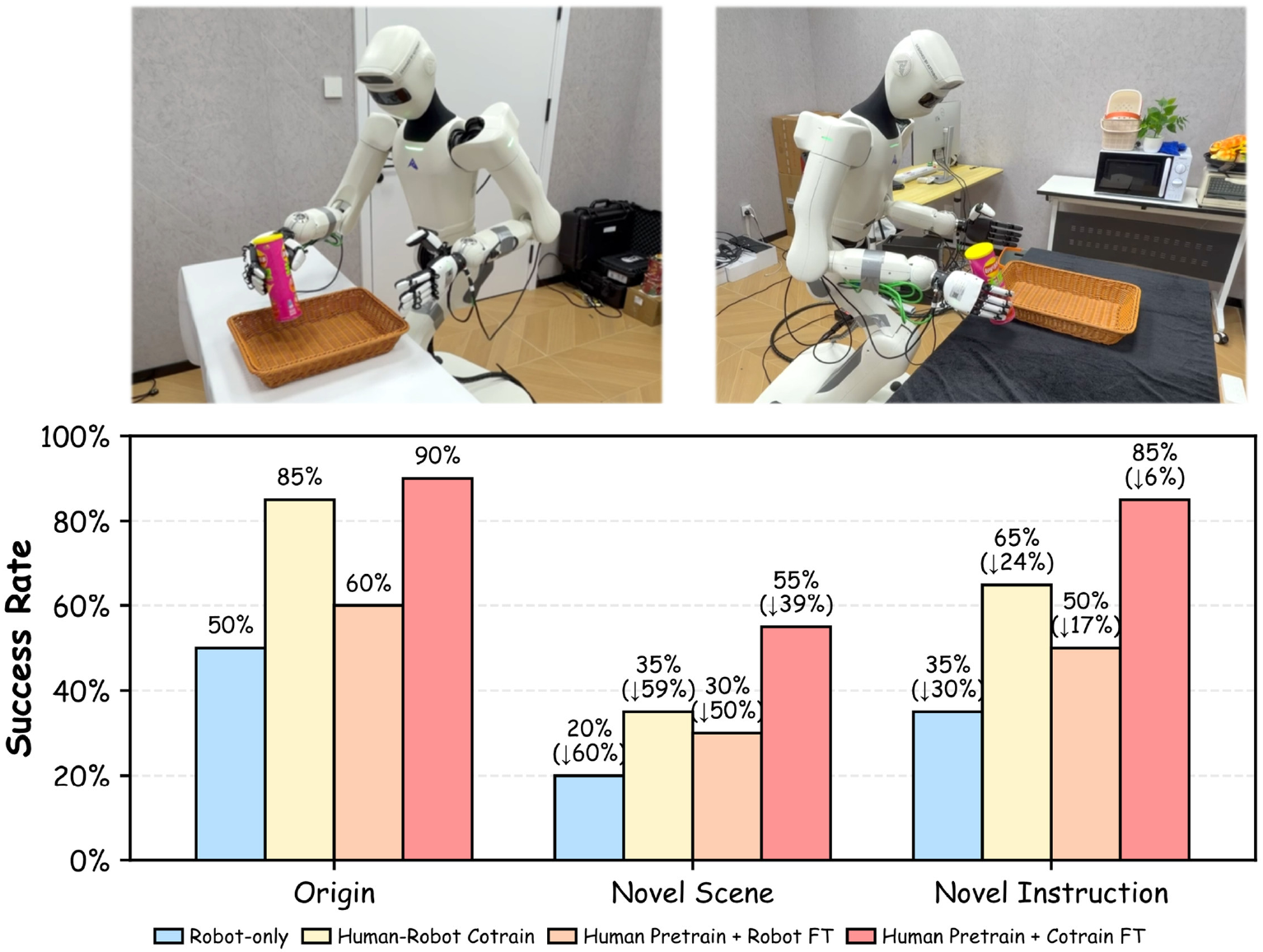}
    \vspace{-0.6cm}
    \caption{\textbf{Generalization under distribution shift.} Evaluation on Pick Chips Can under the original setting, a novel scene, and an unseen instruction phrasing. The images above show the original and novel scenes. Human-assisted training degrades far less than Robot-only under both shifts, especially when human data is retained during finetuning.}
\label{fig:generalization}
\vspace{-0.4cm}
\end{figure}

We additionally evaluate two distribution shifts on Pick Chips Can — a changed background scene and an unseen instruction phrasing. As shown in Fig.~\ref{fig:generalization}, we observe consistent improvements under both background and instruction shifts.
For background shift, Robot-only drops from 50\% to 20\%, while Human Pretrain + Cotrain FT decreases from 90\% to 55\%, maintaining the highest robustness in the novel scene.
For instruction shift, strategies that retain human data during finetuning show greater robustness, with Cotrain and Human Pretrain + Cotrain FT exhibiting 24\% and 6\% relative declines, respectively.
These results suggest that aligned human supervision improves robustness beyond the training distribution, particularly when human data continues to participate during policy optimization.

\section{Related Work}
\subsection{Mobile Manipulation}

Embodied intelligence increasingly seeks generalist models that integrate multimodal perception, language understanding, world modeling, and action generation to accomplish diverse tasks in the physical world~\cite{black2026pi0visionlanguageactionflowmodel,intelligence2025pi05visionlanguageactionmodelopenworld,intelligence2025pi06vlalearnsexperience,gr00tn1_2025,ji-etal-2025-pku,10.1609/aaai.v40i8.37595,li2026wam4dfast4dworld,xu2026twinrldigitaltwindrivenreinforcement,lou2026dream,chi2025wowworldomniscientworld,jia2025video2act,fan2026wowwovalcomprehensive,chi2025mindlearningdualsystemworld,zhang2025worldmodelsbenefitvlms}. Mobile manipulation requires navigation, whole-body reachability, and object interaction to be coordinated over spatially extended tasks.
Accessible platforms and data systems have progressed from Stretch and Mobile ALOHA to integrated or modular whole-body pipelines such as TidyBot++, BRS, TeleMoMa, MoMa-Teleop, HoMeR, and SuperSuit~\cite{kemp2022designstretchcompactlightweight,fu2024mobilealohalearningbimanual,wu2024tidybotopensourceholonomic,jiang2025behaviorrobotsuitestreamlining,dass2024telemomamodularversatile,honerkamp2025wholebodyteleoperationmobile,sundaresan2025homerlearninginthewild,chen2026supersuitisomorphicbimodal}. Humanoid systems including HumanPlus, HOMIE, TWIST/TWIST2, HumanoidExo, and Teleopit further extend teleoperation and policy learning to locomotion, posture, active vision, and dexterous hands~\cite{fu2024humanplushumanoidshadowingimitation,ben2025homiehumanoidlocomanipulationisomorphic,ze2025twistteleoperatedwholebodyimitation,ze2025twist2scalableportable,zhong2025humanoidexoscalablewholebodyhumanoid,wu2026teleopitfullembodiment}. Beyond collection interfaces, generalist and whole-body VLA or world-action models such as GR00T N1, PanoVLA, $\Psi_0$, OpenHLM, and $\omega$-0 increasingly address language-conditioned bimanual, wheeled, and humanoid loco-manipulation~\cite{gr00tn1_2025,yang2026panoramaawarevla,wei2026psi0openfoundationmodel,hu2026openhlmempiricalrecipewholebody,li2026omega0latentpredictive}. HERMES also targets mobile bimanual dexterity but connects learned manipulation skills to a separate navigation module~\cite{yuan2025hermeshumanrobot}. Despite this progress, many systems still require robot-in-the-loop collection, specialized interfaces, or substantial embodiment-specific robot data, motivating scalable robot-free demonstrations for continuous whole-body learning.

\subsection{Human-Robot Alignment}

Transferring human demonstrations to robots requires resolving differences in visual observations, morphology, action semantics, dynamics, and execution speed. Existing work learns cross-embodiment visual or task representations~\cite{zakka2021xirlcrossembodiment,nair2022r3muniversalvisualrepresentation,ma2023vipuniversalvisualreward}, expresses behavior through transferable affordances, flow, or point trajectories~\cite{bahl2023affordanceshumanvideosversatile,yuan2024generalflowfoundationaffordance,ren2025motiontracksunifiedrepresentation,han2026a4acrossembodiment}, or translates and retargets human data into robot-compatible observations and actions~\cite{xie2025human2robotlearningrobotactions,li2025mimicdreameraligninghumanrobot,wang2026ego2robotscalabledata,jeong2026hurorobotizinghumanvideos,araujo2025retargetingmattersgeneralmotion,chen2026warpwholebodyretargetinglearning,lin2026humanashumanoid,xia2026morphologyawarehumanmotion}. For mobile whole-body learning, recent systems often make transfer tractable through discrete locomotion commands, phase-aware policies, decoupled base and manipulation actions, or end-effector and keypoint targets executed by downstream controllers~\cite{shi2026egohumanoidunlockinginthewildlocomanipulation,zhu2025emmascalingmobilemanipulation,xu2026hommilearningwholebodymobile,nai2026humanoidmanipulationinterfacehumanoid,zhao2026halomilearninghumanoidlocomanipulation,wang2026bifrostumibridgingrobotfreedemonstrations,huang2026mobileumicrossview}. These abstractions improve compatibility but may omit part of the original coupled motion. DexRoam instead explicitly aligns embodiment, action semantics, and task progress while retaining a continuous, temporally uninterrupted, jointly predicted base--torso--arm--head--hand action space.

\subsection{Human Demonstrations for Robot Learning}

Human demonstrations provide diverse behavioral priors at substantially lower collection cost than robot teleoperation.~\cite{ye2026datapyramidembodiedmanipulation} Large egocentric datasets have progressed from first-person activities and hand--object interactions to paired viewpoints, metric geometry, full-body motion, and robot-oriented dexterous trajectories~\cite{damen2020epickitchensdatasetcollectionchallenges,grauman2022ego4dworld3000hours,liu2024hoi4d4degocentricdataset,grauman2023egoexo4d,ma2024nymeriamassivecollection,banerjee2024hot3dhandobject,hoque2026egodexlearningdexterousmanipulation,punamiya2026egoverseegocentrichuman,li2026openaoeopenegocentric,lepert2026phantomtrainingrobotsrobots}. Early work mainly extracted reusable representations or rewards, whereas portable interfaces and transfer pipelines now convert human experience into policy supervision for fixed-base, bimanual, and dexterous manipulation~\cite{nair2022r3muniversalvisualrepresentation,ma2023vipuniversalvisualreward,chi2024universalmanipulationinterfaceinthewild,chen2024arcapcollectinghighqualityhuman,kareer2024egomimicscalingimitationlearning,liu2025egozerorobotlearningsmart,wang2024dexcapscalableportablemocap,cheng2024opentelevisionteleoperationimmersiveactive,xu2025dexumiusinghumanhand,yuan2025motiontranshumanvrdata,tao2026dexwilddexteroushumaninteractions,bi2025hrdthumanmanipulationenhanced,yang2025egovlalearningvisionlanguageactionmodels,zheng2026egoscalescalingdexterousmanipulation,liu2026egoengineegocentrichumanvideos,yu2026seedumisharingexoskeleton}. Human-centric VLA and world-model pretraining further scale this idea across data sources and embodiments~\cite{luo2025beingh0visionlanguageactionpretraininglargescale,luo2026beingh05scalinghumancentricrobot,luo2026beingh07latentworldactionmodel,cai2025innonscalingegocentricmanipulation,li2026egowamworldactionmodels}. Complementary to these learning-based representations, recent works also investigate 3D-aware scene modeling for embodied perception~\cite{lepert2026phantomtrainingrobotsrobots,liu2026egoengineegocentrichumanvideos,lepert2025masqueradelearninginthewildhuman,sun2024gsrender,qian2025wristworldgeneratingwristviews4d,xiu2026egotwin,qi2025cocogesturecoherentcospeech3d}. More recently, EMMA, EgoHumanoid, HoMMI, HuMI, BifrostUMI, HALOMI, HERMES, WARP, and OpenHLM extend human supervision to mobile or whole-body policies~\cite{zhu2025emmascalingmobilemanipulation,shi2026egohumanoidunlockinginthewildlocomanipulation,xu2026hommilearningwholebodymobile,nai2026humanoidmanipulationinterfacehumanoid,wang2026bifrostumibridgingrobotfreedemonstrations,zhao2026halomilearninghumanoidlocomanipulation,yuan2025hermeshumanrobot,chen2026warpwholebodyretargetinglearning,hu2026openhlmempiricalrecipewholebody}. Preserving mobile, bimanual, and finger-level coordination in one robot-native policy interface nevertheless remains comparatively underexplored.

\section{Conclusion}

We presented \textbf{DexRoam}, a complete system for learning mobile bimanual dexterous manipulation from egocentric whole-body human demonstrations. Rather than simplifying human motion to ease transfer, DexRoam preserves its continuous, coupled structure end to end. Whole-body motion and egocentric video are captured with only a consumer VR headset and a head-mounted stereo camera. Three explicit alignment stages---embodiment, action-semantic, and temporal---then map this motion into a unified whole-body robot action space, so that VLA policies can consume human and robot data jointly without human-specific action heads.

On a real mobile bimanual dexterous robot, transferred trajectories are directly executable, alignment substantially closes the human--robot action distribution gap, and aligned human demonstrations consistently improve policy learning across five tasks and two VLA backbones while halving the required robot demonstrations. Ablations confirm that each alignment stage is necessary.

\section{Limitations}
Our hands are position-controlled without force or tactile feedback,
so commands cannot encode the closing force a firm grasp requires; combined with execution
error, this yields grasps that touch but do not secure the object. Supervising hand
commands from the Manus glove joint angles, rather than the robot hand state which
saturates at contact, would partially help, while force or tactile feedback is a more
fundamental remedy.
\section*{Acknowledgments}
This work is supported in part by the National Natural Science Foundation of China (NSFC) under Grant No. T2600237, the Early Career Scheme (ECS) of the Research Grants Council (RGC) of the Hong Kong Special Administrative Region, China, under Project No. 26601126, the research funding under the HKUST-DXM AI for Finance Joint Laboratory (DXM25EG01), the National Natural Science Foundation of China under Grant No. 62476011, and the Beijing Natural Science Foundation under Grant No. L252060.

\bibliographystyle{IEEEtran}
\bibliography{references}

\clearpage
\appendix

\subsection{System and Implementation Details}
\label{app:system_details}

This section provides additional implementation details of the DexRoam data acquisition pipeline.
The robot platform, whole-body teleoperation interface, and tracker-free human demonstration system are described in the main paper.
Here, we focus on the actual data rates, synchronization procedure, and visual preprocessing used in our implementation.

\subsubsection{Robot Teleoperation and Data Acquisition}
\label{app:robot_acquisition}

Robot demonstrations are collected using the Astribot--XHand whole-body teleoperation system.
The system records egocentric stereo RGB observations, robot joint states, MANUS hand poses, and XHand joint feedback.
These data streams run asynchronously at their native rates.
In practice, the head stereo RGB stream operates at approximately 30~Hz, the robot joint states at approximately 250~Hz, and both the MANUS hand poses and XHand joint feedback at approximately 39~Hz.
These values reflect the actual data rates observed during data collection rather than the target frequencies specified in the software configuration.

During offline preprocessing, we use the timestamps of the head visual stream as the reference timeline and align the remaining higher-frequency or asynchronous state streams to the corresponding visual frames.
After synchronization, each training sample contains the visual observation, robot state, and action information associated with the same reference time, resulting in robot demonstrations organized at approximately 30~Hz.

The robot head camera stores stereo RGB observations as side-by-side JPEG images with a resolution of $1280\times480$.
Each frame is split at the horizontal midpoint into left and right views, each with a resolution of $640\times480$, which are used as the stereo visual inputs to the policy.

\subsubsection{Human Egocentric Capture System}
\label{app:human_capture}

The tracker-free human demonstration system consists of a Meta Quest 3 and a head-mounted ZED Mini camera. The ZED Mini records stereo RGB observations at approximately 30~Hz, while head and body poses are sampled within the same camera-driven acquisition loop and stored at the same rate. At each iteration, the latest head and body poses are read first, followed immediately by the corresponding ZED image capture, and the pose and stereo image are paired according to their capture order. The visual stream displayed inside the VR headset runs at approximately 20~Hz and is used only for real-time feedback to the demonstrator without affecting the recorded data frequency. Egocentric RGB observations are captured using the ZED Mini in \texttt{HD1080} mode at 30~fps and stored as side-by-side stereo images, with each view at approximately $1920\times1080$ and the full stereo frame at approximately $3840\times1080$. During preprocessing, each frame is decoded and converted to RGB, split into left and right views, center-cropped to $1280\times720$ per view, and resized to $640\times480$. The resulting left and right images are used as the stereo visual inputs for the human demonstrations.

\subsection{Additional Experimental Results and Analysis}
\label{app:additional_results}

\subsubsection{Detailed Tracker-Based Capture Results}
\label{app:pico_results}

In the main paper, we compare our tracker-free capture system with a tracker-based counterpart using a PICO headset and five body-worn motion trackers, and report the replay errors averaged over 15 trajectories. Here, we provide the detailed results underlying this comparison. For each of the three tasks, we evaluate five captured trajectories and report the positional and rotational replay errors of the torso, two arms, and active head.

\begin{table}[t]
\centering
\caption{Detailed replay errors of the PICO + Motion Trackers capture system. Position and rotation errors are reported in millimeters (mm) and degrees ($^\circ$), respectively.}
\label{tab:pico_detailed}
\setlength{\tabcolsep}{0pt}
\renewcommand{\arraystretch}{1}
\begin{tabular*}{\columnwidth}{@{}l@{\hspace{5pt}}c@{\extracolsep{\fill}}cccc@{}}
\toprule
\multicolumn{1}{c}{Component} & Metric & Pick Chips Can & Pour Water & Throw Trash & Average \\
\midrule
\multirow{2}{*}{Torso} & Position & 14.76 & 11.50 & 13.22 & \textbf{13.16} \\
& Rotation & 1.71 & 2.74 & 1.74 & \textbf{2.06} \\
\midrule
\multirow{2}{*}{Left Arm} & Position & 3.85 & 4.11 & 4.41 & \textbf{4.12} \\
& Rotation & 0.54 & 0.58 & 0.60 & \textbf{0.57} \\
\midrule
\multirow{2}{*}{Right Arm} & Position & 5.44 & 5.45 & 6.02 & \textbf{5.63} \\
& Rotation & 0.79 & 0.79 & 0.85 & \textbf{0.81} \\
\midrule
Head & Rotation & 1.27 & 1.32 & 1.09 & \textbf{1.23} \\
\bottomrule
\end{tabular*}
\end{table}

\begin{table}[t]
\centering
\caption{Motion smoothness of retargeted human trajectories after temporal resampling and teleoperated robot demonstrations. Higher LDLJ and SPARC indicate smoother motion.}
\label{tab:smoothness}
\footnotesize
\setlength{\tabcolsep}{0pt}
\renewcommand{\arraystretch}{1}
\begin{tabular*}{\columnwidth}{@{}l@{\extracolsep{\fill}}lcc@{}}
\toprule
\textbf{Task}
& \textbf{Data}
& \textbf{LDLJ $\uparrow$}
& \textbf{SPARC $\uparrow$} \\
\midrule

\multirow{2}{*}{Pick Chips Can}
& Retargeted Human & -20.2802 & -5.8049 \\
& Teleoperated Robot & -20.4726 & -7.0548 \\
\midrule

\multirow{2}{*}{Pour Water}
& Retargeted Human & -20.5962 & -5.9222 \\
& Teleoperated Robot & -20.4183 & -6.5718 \\
\midrule

\multirow{2}{*}{Throw Trash}
& Retargeted Human & -20.8349 & -6.6556 \\
& Teleoperated Robot & -20.3022 & -6.9229 \\
\midrule

\multirow{2}{*}{Deliver Fruit}
& Retargeted Human & -20.3453 & -5.4803 \\
& Teleoperated Robot & -19.8355 & -6.2193 \\
\midrule

\multirow{2}{*}{Push Chair \& Close Laptop}
& Retargeted Human & -20.5130 & -7.2351 \\
& Teleoperated Robot & -20.5575 & -8.4923 \\
\midrule

\multirow{2}{*}{\textbf{Average}}
& \textbf{Retargeted Human} & \textbf{-20.5139} & \textbf{-6.2196} \\
& \textbf{Teleoperated Robot} & \textbf{-20.3172} & \textbf{-7.0522} \\
\bottomrule
\end{tabular*}
\end{table}

\subsubsection{Motion Smoothness Analysis}
\label{app:smoothness}

We further evaluate whether transferring human motion into the robot action space introduces additional motion jitter. We compare the final retargeted human trajectories with teleoperated robot demonstrations using two standard motion-smoothness metrics: log dimensionless jerk (LDLJ) and spectral arc length (SPARC). 

LDLJ measures temporal irregularity through the third-order derivative (jerk) of a scalar trajectory $x(t)$, computed as
\begin{equation}
\mathrm{DLJ}(x)
\propto
\int_{0}^{T}
\left(
\frac{d^{3}x(t)}{dt^{3}}
\right)^{2}
dt ,
\end{equation}

\begin{equation}
\mathrm{LDLJ}(x)
=
-\ln\left|\mathrm{DLJ}(x)\right|.
\end{equation}
where the jerk term is normalized by movement duration and scale to obtain the dimensionless quantity.

SPARC measures smoothness in the frequency domain from the velocity profile. Let $\hat{V}(\omega)$ denote the normalized Fourier magnitude spectrum of the trajectory velocity. SPARC is defined as the negative arc length of the normalized spectrum,
\begin{equation}
\mathrm{SPARC}
=
-\int_{0}^{\omega_c}
\sqrt{
\left(\frac{1}{\omega_c}\right)^2
+
\left(
\frac{d\hat{V}(\omega)}{d\omega}
\right)^2
}
\,d\omega ,
\end{equation}
where $\omega_c$ is determined using a maximum cutoff frequency of 10~Hz and an amplitude threshold of 0.05. A smoother trajectory has a less complex velocity spectrum and therefore a higher SPARC value.

For each episode, LDLJ and SPARC are averaged across all scalar channels of the torso, left arm, right arm, and active head, and the reported values are then averaged over episodes.
The retargeted human trajectories exhibit motion smoothness comparable to teleoperated robot demonstrations. Their mean LDLJ is close to that of robot trajectories ($-20.51$ vs.\ $-20.32$), while their mean SPARC is higher ($-6.22$ vs.\ $-7.05$). At the task level, the retargeted trajectories achieve higher SPARC on all five evaluated tasks, while LDLJ remains within a similar range to the robot demonstrations. These results indicate that the human-to-robot transfer pipeline preserves the smooth temporal structure of human motion and does not introduce evident additional jitter into the resulting robot-space trajectories.

\begin{figure*}[t]
\centering
\includegraphics[width=0.95\textwidth]{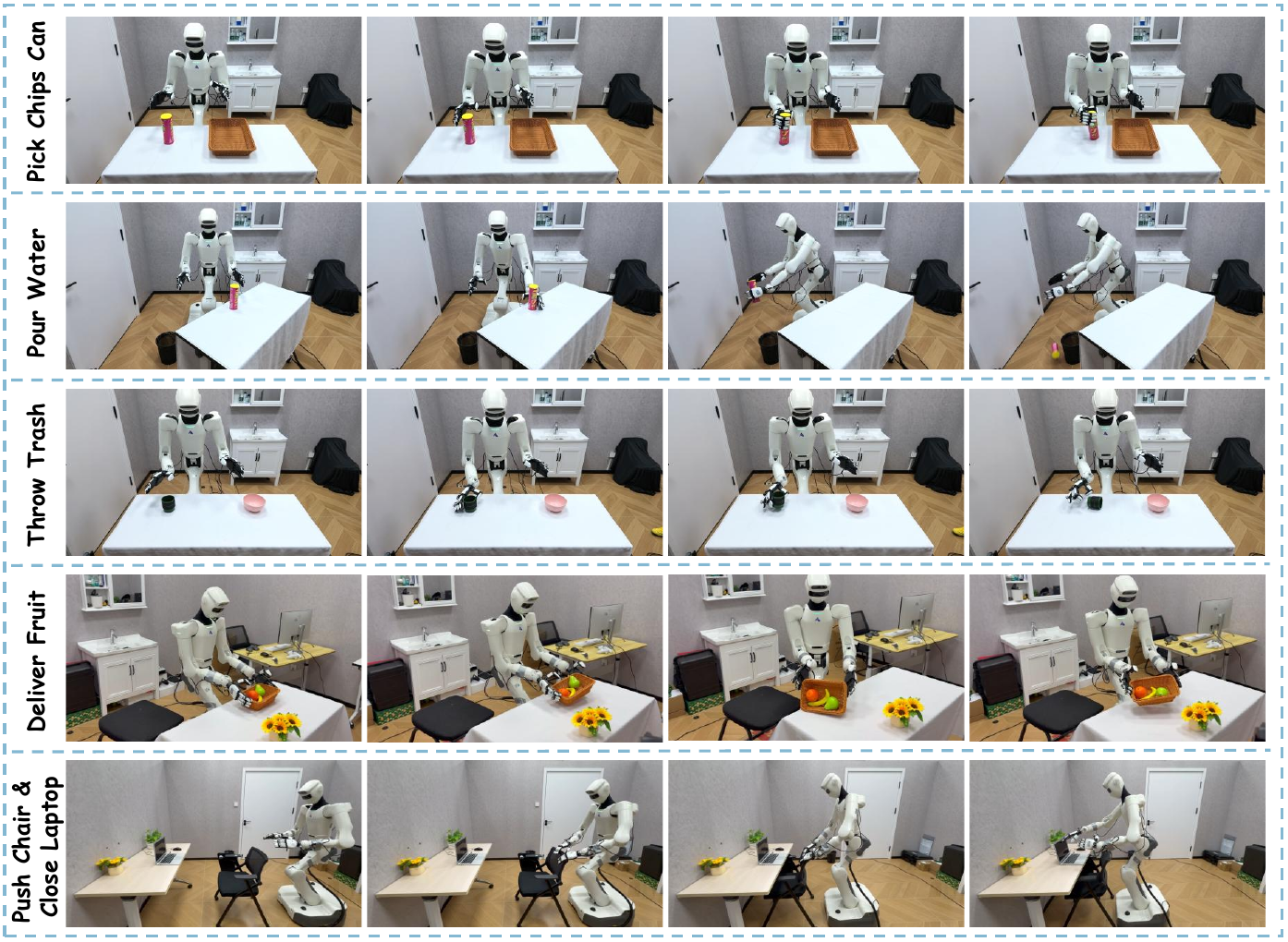}
\caption{Representative failure cases across the five tasks. Each row shows one task,
with successive frames illustrating a failed rollout from left to right.}
\label{fig:failure_cases}
\end{figure*}

\subsection{Failure Mode Analysis}
\label{app:failure}

Across all five mobile dexterous tasks, the robot must coordinate the mobile base, dual-arm
trajectories, head viewpoint, and dexterous-hand joints simultaneously.
Although adding human demonstrations substantially improves overall success rates,
several representative failure modes remain during real-robot deployment, as illustrated
in Fig.~\ref{fig:failure_cases}. We analyze the main failure causes for each task below.

\paragraph{Pick Chips Can}
The dominant failure is the dexterous hand failing to form a stable grasp. The robot can
usually move close to the target and produce a visually plausible reach-and-close motion;
however, in some failures the fingers stop tightening after only light contact with the
can, leaving the grasp insufficiently secure. Although the end-effector pose and hand
shape are close to those of successful demonstrations, inadequate finger closure prevents
the robot from reliably lifting and transporting the can.

\paragraph{Pour Water}
Two typical failures occur. First, the grasped cup may slip during transport or pouring
because a small grasp-position error or insufficient finger closure leaves the grasp
unstable. Second, premature finger contact during the approach can displace the lightweight
cup from its expected position, causing the subsequent closure motion to miss the intended
grasp configuration.

\paragraph{Throw Trash}
Failures mainly arise from timing errors between base motion and object release. Because
the bin opening is relatively small, the robot must release the object at an appropriate
position after approaching the bin. In some failures, the base moves slightly beyond the
desired release position before the action chunk completes, causing the object to miss the
bin. This highlights the need for precise coordination among locomotion, arm motion, and
release timing.

\paragraph{Deliver Fruit}
The main failure arises from insufficient bimanual grasp stability during transport.
In some rollouts, the two hands do not clamp the basket tightly enough, allowing the basket
to oscillate while the robot moves. The resulting motion of the fruit changes the load
distribution and shifts the effective center of mass of the basket, which further increases
its tilt. These effects can accumulate during locomotion and eventually cause the basket
to tip over.

\paragraph{Push Chair \& Close Laptop}
Failures in this long-horizon task often originate from the preceding chair-pushing stage.
In some rollouts, the robot does not push the chair sufficiently close to the table, leaving
the robot too far from the laptop when transitioning to the closing stage. As a result, the
laptop falls outside the effective reaching range of the arm and the robot cannot complete
the final closing motion. This illustrates how errors in an early stage of a long-horizon
mobile manipulation task can propagate and make subsequent manipulation infeasible.

\begin{table}[t]
\centering
\caption{Composition of the 53-D proprioceptive state.}
\label{tab:proprio_state}
\small
\begin{tabular}{lcc}
\toprule
\textbf{Component} & \textbf{Representation} & \textbf{Dim.} \\
\midrule
Left arm    & 3D Cartesian position + 6D rotation & 9 \\
Right arm   & 3D Cartesian position + 6D rotation & 9 \\
Torso       & 3D Cartesian position + 6D rotation & 9 \\
Left XHand  & Joint feedback & 12 \\
Right XHand & Joint feedback & 12 \\
Active head & Joint state & 2 \\
\midrule
\textbf{Total} & & \textbf{53} \\
\bottomrule
\end{tabular}
\end{table}

\begin{table}[t]
\centering
\caption{Composition of the 56-D unified whole-body action.}
\label{tab:whole_body_action}
\small
\begin{tabular}{lcc}
\toprule
\textbf{Component} & \textbf{Representation} & \textbf{Dim.} \\
\midrule
Left XHand   & Joint target & 12 \\
Right XHand  & Joint target & 12 \\
Left arm     & Relative translation + 6D rotation & 9 \\
Right arm    & Relative translation + 6D rotation & 9 \\
Torso        & Relative translation + 6D rotation & 9 \\
Active head  & Joint target & 2 \\
Mobile base  & $(\Delta x,\Delta y,\Delta\theta)$ & 3 \\
\midrule
\textbf{Total} & & \textbf{56} \\
\bottomrule
\end{tabular}
\end{table}

\subsection{Policy Training Details}
\label{app:policy_training}

This section provides additional details of the unified policy interface and the training configurations used for the two VLA backbones.
GR00T N1.7 and $\pi_{0.5}$ use the same stereo visual observations, 53-D proprioceptive state representation, and 56-D whole-body action space.

\subsubsection{Unified Policy Interface}
\label{app:policy_interface}

The policy input at time step $t$ is represented as
\begin{equation}
o_t = \left(I_t^L, I_t^R, p_t, \ell\right),
\end{equation}
where $I_t^L$ and $I_t^R$ denote the left and right egocentric RGB observations, respectively, $p_t$ denotes the robot proprioceptive state, and $\ell$ is the task language instruction.
Both stereo views are represented at a resolution of $640\times480$.
The current implementation uses a single-frame proprioceptive state without explicitly stacking multiple state histories.

\paragraph{Unified Policy State and Action Space}

The robot proprioceptive state is represented as $p_t \in \mathbb{R}^{53}$, and the whole-body action is represented as $a_t \in \mathbb{R}^{56}$. Their compositions are summarized in Tables~\ref{tab:proprio_state} and~\ref{tab:whole_body_action}.

The Cartesian states of the two arms and torso are represented by 3D positions and continuous 6D rotations. The policy state does not explicitly include the chassis pose. For actions, the two arms and torso use relative translation and 6D rotation, the mobile base uses local planar motion $(\Delta x,\Delta y,\Delta\theta)$, and the active head and two XHands use robot joint-space targets.

\begin{table}[t]
\centering
\caption{Training configuration for $\pi_{0.5}$.}
\label{tab:pi05_training}
\small
\begin{tabular*}{0.92\columnwidth}{@{}l@{\extracolsep{\fill}}>{\centering\arraybackslash}p{0.50\columnwidth}@{}}
\toprule
\textbf{Hyperparameter} & \textbf{Value} \\
\midrule
Base model & $\pi_{0.5}$ \\
Number of GPUs & 2 \\
Maximum training steps & 30,000 \\
Global batch size & 4 \\
Per-GPU batch size & 2 \\
Gradient accumulation steps & 1 \\
Optimizer & AdamW \\
Learning rate & $2.5\times10^{-5}$ \\
Weight decay & $1\times10^{-10}$ \\
Warmup ratio & 0.033 \\
Warmup steps & 1,000 \\
\bottomrule
\end{tabular*}
\end{table}

\begin{table}[t]
\centering
\caption{Training configuration for GR00T N1.7.}
\label{tab:groot_training}
\small
\begin{tabular*}{0.92\columnwidth}{@{}l@{\extracolsep{\fill}}>{\centering\arraybackslash}p{0.50\columnwidth}@{}}
\toprule
\textbf{Hyperparameter} & \textbf{Value} \\
\midrule
Base model & NVIDIA Isaac-GR00T N1.7-3B \\
Number of GPUs & 4 \\
Maximum training steps & 40,000 \\
Global batch size & 24 \\
Per-GPU batch size & 6 \\
Gradient accumulation steps & 1 \\
Optimizer & AdamW \\
Learning rate & $1\times10^{-4}$ \\
Weight decay & $1\times10^{-5}$ \\
Warmup ratio & 0.05 \\
Warmup steps & 2,000 \\
\bottomrule
\end{tabular*}
\end{table}

\subsubsection{$\pi_{0.5}$ Implementation}
\label{app:pi05}

For $\pi_{0.5}$, each training sample consists of the left and right RGB observations, the 53-D proprioceptive state, the task language instruction, and the corresponding 56-D whole-body action targets. We initialize the model from the pretrained $\pi_{0.5}$ base checkpoint. Since the pretrained action interface does not match the 56-D DexRoam action space, the action input and output projections are randomly initialized. All model parameters are trainable during finetuning, with no frozen modules.
The complete training configuration is summarized in Table~\ref{tab:pi05_training}.

\subsubsection{GR00T N1.7 Implementation}
\label{app:groot}

For GR00T, we use NVIDIA Isaac-GR00T N1.7-3B as the base model. The model takes stereo egocentric RGB observations, the 53-D proprioceptive state, and the task language instruction, and predicts actions in the same 56-D whole-body action space. During training, the pretrained LLM and vision backbones are frozen, while the projector and diffusion action decoder are optimized to adapt the pretrained model to the DexRoam action interface.
The complete training configuration is summarized in Table~\ref{tab:groot_training}.

\subsection{Detailed Evaluation Protocol and Metrics}
\label{app:evaluation}

This section provides the detailed evaluation protocol for task progress and success, as well as the implementation details of the human--robot action distribution metrics used in the main paper.

\subsubsection{Human--Robot Action Distribution Analysis}
\label{app:distribution_metrics}

We analyze the distribution gap between human-derived and real-robot actions over the complete 56-D whole-body action space, including the two 12-D dexterous-hand actions, two 9-D arm actions, 9-D torso action, 2-D head action, and 3-D mobile-base action.

\paragraph{Action normalization}
Before computing MMD, SWD, and PCA, each action dimension is independently normalized using pooled robust statistics. For the $j$-th action dimension, we use
\begin{equation}
\tilde{a}_j =
\frac{
a_j-\operatorname{median}(a_j)
}{
\max\left(
(Q_{75,j}-Q_{25,j})/1.349,\,
10^{-3}
\right)
},
\label{eq:action_normalization}
\end{equation}
where $Q_{75,j}$ and $Q_{25,j}$ denote the 75th and 25th percentiles, respectively. For each episode, we randomly sample up to 256 frames to estimate the pooled statistics. Each data source is further limited to at most 4,000 frames, and all participating sources are downsampled to the same number of samples before pooling, such that each domain contributes equally. Human and robot actions are normalized in the same coordinate system.

\paragraph{Maximum Mean Discrepancy}
We compute Maximum Mean Discrepancy (MMD) on the normalized 56-D action vectors using a single Gaussian RBF kernel. Given equally sized human and robot action sets $\{x_i\}_{i=1}^{n}$ and $\{y_i\}_{i=1}^{n}$, respectively, we use the biased V-statistic estimator
\begin{equation}
\begin{aligned}
\mathrm{MMD}
=
\Bigg[
\max\Bigg(
&\frac{1}{n^2}\sum_{i,j=1}^{n} k(x_i,x_j)
+\frac{1}{n^2}\sum_{i,j=1}^{n} k(y_i,y_j) \\
&-\frac{2}{n^2}\sum_{i,j=1}^{n} k(x_i,y_j),
\,0
\Bigg)
\Bigg]^{1/2}.
\end{aligned}
\label{eq:mmd}
\end{equation}
where the Gaussian kernel is
\begin{equation}
k(x,y)
=
\exp\left(
-\frac{\|x-y\|_2^2}{2b}
\right).
\label{eq:mmd_kernel}
\end{equation}
The bandwidth parameter $b$ is set to the median pairwise squared Euclidean distance over the pooled normalized samples, corresponding to $\sigma^2=b$ under the standard Gaussian-kernel notation. The bandwidth is estimated separately for the absolute and relative action spaces, while the raw and aligned relative distributions share the same bandwidth. For each comparison, we sample the same number of human and robot actions without replacement, with at most 4,000 samples from each domain.

\paragraph{Sliced Wasserstein Distance}
We use the first-order Sliced Wasserstein Distance ($W_1$) on the normalized 56-D action space. We sample $L=128$ random projection directions from a standard multivariate Gaussian distribution and normalize each direction to unit length. For each direction, the projected human and robot samples are sorted, and the absolute differences between corresponding ordered samples are averaged. Specifically,
\begin{equation}
\mathrm{SWD}
=
\frac{1}{L}
\sum_{\ell=1}^{L}
\frac{1}{n}
\sum_{i=1}^{n}
\left|
x^{(\ell)}_{(i)}
-
y^{(\ell)}_{(i)}
\right|,
\label{eq:swd}
\end{equation}
where $x^{(\ell)}_{(i)}$ and $y^{(\ell)}_{(i)}$ denote the $i$-th ordered samples after projection onto direction $\ell$. Human and robot samples are balanced and limited to at most 4,000 actions per domain. Random projection directions are independently sampled for different alignment stages.

\paragraph{PCA visualization}
PCA is performed on the normalized 56-D action vectors using the same robust normalization described above. To avoid imbalance among data sources, samples are balanced across domains, with at most 6,000 samples used to fit each PCA model. The first two principal components are retained for visualization.

\subsubsection{Task Completion Score}
\label{app:tcs}

We use Task Completion Score (TCS) to characterize partial progress during long-horizon mobile manipulation. Each task is decomposed into a sequence of ordered milestones with task-specific weights. The TCS of a rollout is computed as the sum of the weights of all completed milestones, with the milestone weights for each task summing to 1.0. Detailed scoring criteria are provided in Table~\ref{tab:tcs_criteria}.

\begin{table*}[t]
\centering
\caption{Task Completion Score (TCS) criteria. Each task is decomposed into five ordered milestones with task-specific weights shown in parentheses.}
\label{tab:tcs_criteria}
\small
\setlength{\tabcolsep}{4pt}
\renewcommand{\arraystretch}{1.15}
\begin{tabular}{p{1.9cm} p{2.65cm} p{2.65cm} p{2.65cm} p{2.65cm} p{2.65cm}}
\toprule
\textbf{Task}
& \textbf{Stage 1}
& \textbf{Stage 2}
& \textbf{Stage 3}
& \textbf{Stage 4}
& \textbf{Stage 5} \\
\midrule

Pick Chips Can
& Move forward to the table (0.2)
& Move the hand into the grasping envelope without knocking over the can, with the fingertip center within 3~cm of the can axis (0.2)
& Close the fingers and establish a stable grasp with at least two opposing fingers in contact with the can (0.2)
& Lift the can above the table by at least the basket height and maintain the grasp for at least 3~s without slipping (0.2)
& Successfully place the can into the basket (0.2)
\\

\midrule

Pour Water
& Reach toward the container (0.1)
& Establish a stable grasp on the container (0.2)
& Lift the container at least 5~cm above the table (0.2)
& Move toward the target while maintaining a stable grasp on the container (0.2)
& Rotate the wrist and pour water into the target container (0.3)
\\

\midrule

Throw Trash
& Move the hand into the grasping envelope of the chips can without knocking it over, with the fingertip center within 3~cm of the can axis (0.2)
& Close the fingers and establish a stable grasp with at least two opposing fingers in contact with the object (0.2)
& Move the mobile base until the trash bin is within the reachable workspace (0.2)
& Release the object such that it falls into the trash bin (0.3)
& Complete the task without displacing or knocking over the trash bin (0.1)
\\

\midrule

Deliver Fruit
& Move both hands into the grasping envelope and establish contact with the fruit basket (0.2)
& Lift the fruit basket at least 5~cm above the supporting surface using coordinated bimanual grasping (0.2)
& Maintain the lifted basket for at least 3~s without dropping it (0.2)
& Complete the required locomotion and turning while keeping the basket and its contents stable (0.2)
& Place the fruit basket at the target location (0.2)
\\

\midrule

Push Chair \&
Close Laptop
& Establish contact between the hand and the back of the chair (0.2)
& Successfully push the chair without losing hand--chair contact (0.2)
& Move the chair to the specified target position under the table (0.2)
& Extend the right hand and successfully reach behind the laptop lid (0.2)
& Successfully close the laptop (0.2)
\\

\bottomrule
\end{tabular}
\end{table*}

\vfill

\end{document}